\documentclass[11pt]{article}

\usepackage[final]{acl}

\usepackage{times}
\usepackage{latexsym}

\usepackage{algorithm}
\usepackage{algorithmic}
\usepackage{amsmath}
\usepackage{amssymb}

\usepackage[T1]{fontenc}

\usepackage[utf8]{inputenc}

\usepackage{microtype}

\usepackage{inconsolata}

\usepackage{graphicx}

\usepackage{comment}
\usepackage{enumitem}
\setlist[description]{style=unboxed,leftmargin=0em,labelsep=0.5em}

\title{Calibrating Model-Based Evaluation Metrics for Summarization}

\author{
Hongye Liu, Dhanajit Brahma, Ricardo Henao \\
Duke University \\
\texttt{\{hongye.liu, dhanajit.brahma, ricardo.henao\}@duke.edu}
}

\usepackage[utf8]{inputenc} 
\usepackage[T1]{fontenc}    
\usepackage{hyperref}        
\usepackage{url}  
\usepackage{booktabs}        
\usepackage{amsfonts}        
\usepackage{nicefrac}        
\usepackage{microtype}        
\usepackage{xcolor}     

\usepackage{amsmath}
\usepackage{graphicx}
\usepackage{float}    

\usepackage{multirow}
\usepackage{booktabs}
\usepackage{arydshln}  
\usepackage[table,xcdraw]{xcolor}
\usepackage{wrapfig}

\newcommand{\ECEc}{$\text{ECE}'$}
\newcommand{\GECEc}{$\text{GECE}'$}
\newcommand{\Brierc}{$\text{Brier}'$}
\newcommand{\Slopec}{$\text{Slope}'$}

\usepackage{enumitem}
\setlist[itemize,1]{leftmargin=1em}

\setcitestyle{round}
\newcommand{\cgt}{t}

\begin{document}
\maketitle

\begin{abstract}
Recent advances in summary evaluation are based on model-based metrics to assess quality dimensions, such as completeness, conciseness, and faithfulness. 
However, these methods often require large language models, and predicted scores are frequently miscalibrated, limiting their reliability.
Moreover, evaluating the average quality across different summaries for a single document typically requires access to multiple reference summaries.
Here, we propose a general framework that generates individual and average proxy scores without relying on reference summaries, human annotations, or expensive model-based metrics.
We also propose {\em group isotonic regression binning} (GIRB)\footnote{The source code is available and can be accessed at \href{ https://github.com/Hyfred/Calibrating-Evaluation-Metrics}{https://github.com/Hyfred/Calibrating-Evaluation-Metrics}.}, a calibration method that adjusts the raw predictions to better align with ground-truth evaluation metrics.
While we focus on continuous-value scenarios, such as summarization, the method is applicable to discrete-value tasks, such as question answering.
Experiments on seven datasets demonstrate that our approach consistently outperforms existing baselines. 
\end{abstract}

\section{Introduction}
Large language models (LLMs) have demonstrated remarkable capabilities in various natural language processing tasks, such as summarization and question answering (QA) \citep{minaee2024large, luo-etal-2025-dtcrs, luo2026gcotdecodingunlockingdeepreasoning}.
However, these models are prone to issues such as hallucination, which results in incorrect or misleading outputs \citep{huang2023survey}.
Consequently, evaluating the outputs of LLMs is critical for their effective use \citep{lan2025f2bench, lan2025mcbe, wang2026one, wang2026creativebench}.
One direct approach to evaluation is through human judgment, but this method is costly and time-consuming \citep{zhang2021finding, liu2025learning}.

To address this issue, {\em model-based evaluation metrics} have been proposed.
For example, FineSurE \citep{song2024finesure} evaluates completeness, conciseness, and confidence, while UniEval \citep{lee2024unisumeval} measures consistency, fluency, and coherence.
Other more general metrics, such as BERTScore \citep{zhang2019bertscore} or BARTScore \citep{yuan2021bartscore}, can also be used.
However, summarization models typically do not produce evaluation scores, which requires additional model-based scoring methods.
These methods often rely on LLMs, making evaluation computationally expensive.
Furthermore, deep neural networks, including LLMs, are known to suffer from calibration issues \citep{guo2017calibration, parappan2026labels}, a limitation that LLM-based evaluation methods also inherit \citep{manggala2024qa}.

To address these challenges, our goal is to develop a simple approach that enables predictors to generate proxy scores calibrated against model-based scores.
For example, under a well-calibrated model, if we collect a set of documents paired with summaries that receive a predicted conciseness score of 0.85, their average model-based conciseness score will also be 0.85.
To this end, \citet{manggala2024qa} proposed a group-wise calibration method for generative question answering systems, aiming to calibrate the confidence score of a question-answer pair to its final binary label, {\em i.e.}, whether the answer is correct.
Their approach first clusters samples in the embedding space to perform adaptive binning, with the goal that each partition contains semantically similar question answer pairs that are ``close'' to each other in the embedding space.
This design is motivated by the limitations of global calibrators, which can obscure important differences between content types.
For example, two question-answer pairs may have similar raw model confidence scores ({\em e.g.}, 0.7), but one involves a factual geography question and the other a medical question; thus, a single global adjustment could miscalibrate one of them, whereas group-specific calibration can potentially correct for such semantic differences.
Within each cluster, the method bins the data by sorting its probabilistic predictions and then assigning the mean true label within each bin as the calibrated output.
Although simple, this approach has the limitation that replacing all values in a bin with the mean creates a piecewise-constant function that may lose fine-grained information.
This limitation suggests that alternative calibration strategies could be explored.

Moreover, \citet{manggala2024qa} focus on calibrating discrete labels, such as binary correctness in QA.
In contrast, tasks like summarization involve continuous metrics, {\em e.g.}, conciseness, where the goal is not merely to indicate whether a summary is concise, but to quantify how concise it is.
However, in this context, the calibration for continuous metrics remains largely unexplored.
In addition, if we wish to determine whether the summary of a document is above average relative to a set of summaries, a straightforward approach would involve generating multiple summary pairs and computing the average score across them.
Here, the average is taken over summaries produced by different summarization systems.
By comparing the score of the current summary with this average, we can assess whether it is better than average.
However, generating multiple summaries and evaluating them repeatedly is computationally expensive.
Therefore, beyond assessing the quality of a single summary, our goal also includes estimating both the score of an individual summary and the average score across multiple summaries in a single pass.
This allows one to evaluate whether a given summary outperforms the average without accessing all reference summaries and to decide whether generating a higher scoring alternative is necessary.

We make the following contributions.
\begin{enumerate}[leftmargin=5mm,topsep=0mm,itemsep=-2mm]
    \item We propose a general framework that generates both individual and reliable \textit{average} proxy scores in document summarization \textit{in a single pass}, without needing reference summaries, human annotations, or expensive model-based metrics.
    Instead of iterating, the method directly produces reliable average scores, eliminating the need for multiple evaluation rounds.
    \item We introduce {\em group isotonic regression binning} (GIRB), which performs group-level calibration to capture bias-causing semantic heterogeneity. Unlike traditional individual-based methods, GIRB adjusts the predictions based on group-wise monotonic trends, improving the consistency between predicted and true metrics.
    \item Although our main focus is on summarization and other continuous-value tasks, the proposed framework is also applicable to discrete-value tasks like QA.
    Experiments across multiple pre-trained models show that our method consistently outperforms existing baselines.
    We test our approach on both summarization and QA tasks, and find that, compared to other calibration methods, GIRB performs better across a variety of calibration error metrics.
\end{enumerate}

\section{Related Work}
\noindent\textbf{Summary Evaluation.}
Traditional summarization metrics, such as ROUGE, have long been used to evaluate summary quality.
However, these metrics are often criticized for their poor correlation with human judgments on semantic fidelity and factuality.
Recent advances have seen the emergence of LLM-based evaluators, which assess summary content at a finer granularity~\citep{laban2023summedits,lu2023toward}. 
For example, FineSurE~\citep{song2024finesure} extracts atomic \emph{keyfacts} from both the source and the summary, computing completeness and conciseness by aligning these keyfacts with the sentences in the summary.
This results in interpretable dimension-specific scores.
Meanwhile, UniSumEval~\citep{lee2024unisumeval} provides a unified and robust framework for benchmarking summarization systems that improves the precision and comprehensiveness of performance evaluations.
Collectively, these works signal a shift from token-level to content-level evaluation, marking a significant step toward more reliable and actionable supervision in summarization.
Our work builds on this trend by leveraging these evaluative scores as a ground truth signal for model calibration.

\noindent\textbf{Calibration for Language Models.}
The calibration of LLMs has been a subject of growing interest, especially as LLMs continue to show remarkable capabilities in various tasks.
However, LLMs often produce outputs with poorly calibrated confidence levels, which can be problematic for tasks that require accurate probabilistic estimates.
The objective in reinforcement learning from human feedback (RLHF), {\em e.g.}, tends to prioritize adherence to user instructions in dialog over the need for well-calibrated predictions~\citep{kadavath2022language}.

\citet{liu2024calibrating, winata2024metametrics} aim to develop evaluation metrics that better correlate with human judgments. \citet{liu2024calibrating} improves alignment by modifying evaluation prompts during score generation, while \citet{winata2024metametrics} aggregates multiple metrics using a boosting-based approach. The latter assumes the availability of several metric scores, which leads to an increase in computational cost. Both works primarily assess performance using rank-based correlation measures such as Spearman’s and Kendall’s coefficients.
Although both works use the term calibration, they do not adopt calibration in a formal statistical sense. Formal notions of calibration have been established in earlier work, including \citet{dawid1982well, degroot1983comparison}, as well as in recent studies on confidence calibration for LLMs.

\citet{lin2022teaching} introduced the concept of verbalized confidence, which prompts LLMs to express their confidence directly, focusing on fine-tuning rather than zero-shot verbalized confidence.
In contrast, \citet{mielke2022reducing} employed an external calibrator for white-box LLMs to adjust confidence levels.
Other approaches focused on consistency measures to improve calibration~\citep{lyu2024calibrating}.
The method proposed by \citet{manggala2024qa} introduced a {\em post hoc} calibration technique that provides stronger calibration guarantees compared to previous approaches.

It is worth noting that prior work has largely studied calibration in discrete settings, such as question answering, where predictions are calibrated against binary labels, or used the term calibration without a
formal definition. 
In contrast, we are the first to provide a formal definition of calibration tailored to the summarization task, together with a systematic study of calibration for summarization evaluation, where the target labels are continuous variables.

\begin{figure*}[t]
    \centering
    \includegraphics[width=\textwidth]{figures/model.png}
    \caption{
    \textbf{Workflow Overview.} We input a document and its summary to produce two types of scores. The average score, reflecting the mean level across multiple summarization systems, serves as a reference. Comparing an individual score with this reference indicates summary quality: above average is good, below average is poor.
    }
    \label{fig:workflow}
\end{figure*}

\vspace{-1mm}
\section{Methodology}
\begin{figure}[t]
    \centering
    \includegraphics[width=0.47\textwidth]{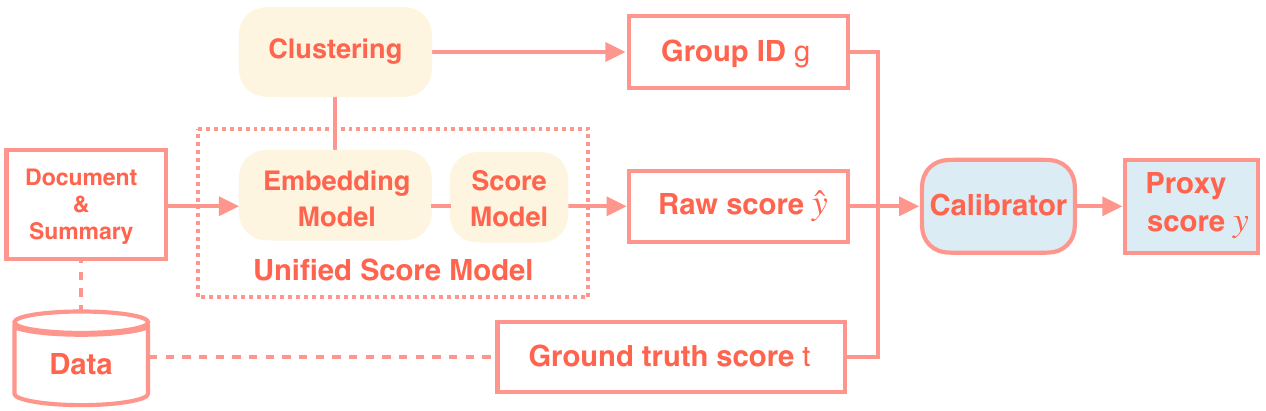}
    \caption{
    \textbf{Framework Overview.} 
    A unified scoring model maps each document-summary pair to raw scores \( \hat{y} \) via a shared embedding used for grouping. GIRB has two steps: $i$) embedding-space clustering to assign a group ID, and $ii$) {\em post hoc} calibration using the group ID, raw scores, and model-based ground truth.
    }
    \label{fig:framework}
\end{figure}

\vspace{-1mm}
In this section, we present our framework for generating calibrated proxy scores for summary evaluation.
We first formalize the problem of predicting both individual and average quality scores across multiple evaluation dimensions without requiring reference summaries or expensive model-based metrics at inference time.
We also motivate a {\em post hoc} calibration approach that produces reliable scores from a \emph{lightweight} calibrator that preserves scalability.
We then describe our method, group isotonic regression binning (GIRB), which calibrates the scores obtained from a LLM by performing clustering and group-based isotonic regression for {\em post hoc} calibration.
Figure~\ref{fig:framework} provides an overview of the complete framework.

\subsection{Problem Definition}
Let $d \in \mathcal{D}$ be a source document, $s \in \mathcal{S}$ a generated summary, and for each document $d$ we collect summaries $\{s^{(1)},\ldots,s^{(N)}\}$ from different systems or LLMs.
For each document-summary pair $(d, s^{(i)})$, we seek to evaluate each summary along $K$ dimensions, {\em e.g.}, completeness, conciseness, faithfulness, {\em etc.}
A direct way to do this is to use human judgments, which is expensive and time-consuming.
To avoid this issue, we use model-based estimation.

Formally, let a scoring model produce a score $\cgt_{k}^{(i)}$ for the summary $s^{(i)}$ on dimension $k$.
We treat this score as the model-based ground-truth metric. 
For example, we can use FineSurE~\citep{song2024finesure} for completeness, conciseness, and faithfulness, and UniEval~\citep{lee2024unisumeval} for consistency, fluency, and coherence.
Other metrics such as BERTScore~\citep{zhang2019bertscore} or BARTScore~\citep{yuan2021bartscore} can also be used.

However, the problem is that the summarization models themselves do not produce evaluation scores and computing model-based scores can also be expensive, since they often rely on LLMs.
Furthermore, deep neural networks are known to be miscalibrated~\citep{guo2017calibration, parappan2026labels}, and LLMs also inherit this limitation.
Our goal is to design a simpler way to make a predictor $h_k$ to generate a proxy score $y_{k}^{(i)}$ that is \emph{calibrated} relative to the model-based ground-truth score $\cgt_{k}^{(i)}$. 
For example, if a model is \emph{perfectly} calibrated, of all summary-document pairs for which \(y_{k}^{(i)}=0.85\), their average FineSure score \(\cgt_{k}^{(i)}\) should also be \(0.85\).
We use $T, Y$ to denote the random variables corresponding to the model-based ground-truth score $t$ and the proxy score $y$, respectively.
Formally, a predictor \(M:\mathcal{D}\times\mathcal{S}\to[0,1]\) is calibrated if for all \(p\in[0,1]\), \(\mathbb{E}[T \mid Y=p]=p\).

Importantly, we want to measure not only the quality of an individual summary but also whether it is better than the average summary quality.
Given $N$ summaries corresponding to one document, we can estimate both the individual score $\cgt_{k}^{(i)}$ of the individual summary $s^{(i)}$ and the average score $\bar{\cgt}_{k}$ for all $N$ summaries.
This allows one to determine whether a single summary is above or below average without needing access to all $N$ summaries.
Figure~\ref{fig:workflow} illustrates judging the quality of an individual summary using the average score as a reference.

During inference, given a document-summary pair $(d,s^{(i)})$, our goal is to produce two types of proxy scores for $K$ evaluation dimensions:
the individual proxy score $y_{k}^{(i)}$ and
the average proxy score $\bar{y}_{k}$, without requiring access to multiple summaries.
Because there are $K$ dimensions, this yields $2K$ target values per $(d,s^{(i)})$.
This approach does not require multiple reference summaries, model-based metrics, or human annotations.
The proxy scores serve as reliable stand-ins for true evaluation, providing both an estimate of the summary's quality and its relative standing compared to the expected average quality.
This enables efficient and scalable evaluation of summaries in real-world settings.






\subsection{Score Modeling and GIRB Calibration}
\noindent{\bf Overview.}
For each document-summary pair $(d, s^{(i)})$ we can obtain model-based scores along $K$ dimensions, which produce $2K$ values: one set of $K$ individual scores $\{t_k^{(i)}\}_{k=1}^K$ and another set of $K$ average scores $\{\bar{t}_k\}_{k=1}^K$. 
We propose the general framework in Figure~\ref{fig:workflow} that generates both individual and \textit{average} proxy scores in a \textit{single pass}, without relying on reference summaries.
Our objective is to train a model that, given only a single document and summary, can predict these $2K$ proxy scores. 
For simplicity, let $\hat{\mathbf{y}} \in \mathbb{R}^{2K}$ denote the raw multi-output predictions for one pair $(d,s)$, where each component corresponds to either an individual or an average target for a specific dimension.
Specifically, we introduce GIRB (groupwise isotonic regression with binning) as the combination of $i$) grouping through clustering on embeddings and $ii$) {\em post hoc} group-wise isotonic calibration applied to raw scores.
GIRB calibrates the predictions of any score model, rather than being part of a unified score model itself.

\noindent{\bf Unified score model (embedding + score model).}
We treat the embedding encoder and score model as a single unified score model:
\vspace{-0.25em}
\begin{equation*}
    \hat{\mathbf{y}} = f_{\theta}(E_{\psi}(d,s)) = M_\Theta(d,s),
\vspace{-0.25em}
\end{equation*}
where $E_{\psi}(\cdot)$ is a pretrained or lightly fine-tuned embedding encoder ({\em e.g.}, a frozen LLM or sentence encoder), $f_{\theta}(\cdot)$ is a multi-output score model, and we use $M_\Theta(\cdot)$ to denote the unified score model.
For ease of explanation, we describe them separately:
{\em Embedding Model:} $E_{\psi}(d,s) \mapsto \mathbb{R}^{m}$, which produces a shared representation that is reused for both scoring and grouping.
{\em Score Model:} A multi-output score model \(f_{\theta}:\mathbb{R}^{m}\to\mathbb{R}^{2K}\), which can be instantiated with any function class; from simple linear mappings to expressive architectures (linear regression, MLP or transformer adapters).

Note that GIRB is not part of \(M_{\Theta}(\cdot)\).
It is a separate downstream model that post-calibrates its outputs (or that of any other score model) without changing its architecture or training objective.




\noindent{\bf Grouping via clustering on embeddings.}
We partition the embedding space into multiple clusters, grouping $(d, s)$ pairs with similar semantic and structural properties.
We define a fixed mapping $\phi:\mathcal{D}\times\mathcal{S}\to\mathcal{T}$ as $g=\phi(d,s)$, where $\mathcal{T}$ indexes clusters obtained by applying a clustering algorithm ({\em e.g.}, KMeans, KD-Tree) to $E_{\psi}(d,s)$.
Thus, each $(d, s)$ pair is assigned a group identifier $g = \phi(d, s)$.
The same embedding feeding the score model is used to assign groups, which ties modeling and calibration while keeping components modular.
This grouping step is the first component of GIRB: it partitions the data into nearby regions in the embedding space, so each region can obtain its own simple, group-specific calibrator.
We introduce clustering because the mapping from raw to calibrated score can differ across content types; thus a single global calibrator can over-correct some cases and under-correct others, whereas per-group mappings are poised to better match local patterns.
Reusing the scoring embeddings for grouping keeps the system simple and avoids additional passes.




\noindent{\bf {\em Post hoc} group calibration.}
Raw scores from a trained unified score model $M_\Theta(\cdot)$ can be miscalibrated, which means that predicted values may not correspond to the same expected ground truth value over the range of predictions; therefore, {\em post hoc} calibration aims to correct it without retraining $M_\Theta(\cdot)$.
This calibration step is the second component of GIRB, namely within each group, we learn a mapping from the raw scores to the target scores so that the model predictions are better calibrated.

The {\em post hoc} calibration algorithm in~\citet{manggala2024qa} first sorts the prediction-target pairs according to their predicted values.
Then it divides the sorted pairs into bins of equal size and uses the mean of the corresponding target values in each bin as the new calibrated value.
This approach relies on a tunable hyperparameter, which is the number of points per bin, and produces a (non-smooth) piecewise constant function.
The imputation of raw predictions with these bin-wise averages ensures that the predicted values match the observed outcomes on average within each bin. 
Consequently, the calibrated value of any prediction in a bin approximates the conditional expectation of the true target given the original prediction, satisfying the definition of calibration.

Instead of specifying fixed bins, we use isotonic regression~\citep{barlow1972isotonic} to learn a continuous piecewise linear function that increases monotonically relative to the raw scores.
Isotonic regression adaptively merges adjacent violations to enforce monotonicity and thus, does not require a bin-size hyperparameter.
For a group $g$ and dimension $k$, given a set of $n_g$ prediction-target pairs $(\hat{y}_k^{(j)}, t_k^{(j)})_{j=1}^{n_g}$ from a held-out calibration split, isotonic regression first sorts the pairs according to the predicted raw scores $\hat{y}_k^{(j)}$.
Then it applies the pool-adjacent violators algorithm (PAVA) \citep{de2010isotone}, scanning the sequence from left to right to ensure that the targets $t_k^{(j)}$ are monotonic.
Whenever a violation is detected, {\em i.e.}, $t_k^{(j)} > t_k^{(j+1)}$ for a non-decreasing fit, the adjacent points are merged into blocks and replaced with their weighted average.
This process repeats until the entire sequence satisfies the monotonicity constraint.
Then linear interpolation between adjacent blocks can be applied to construct a continuous isotonic function.

During inference, a new prediction score is mapped to the corresponding value on this isotonic function, resulting in the calibrated score.
This means that isotonic regression fits a non-decreasing function \(h:\mathbb{R}\to\mathbb{R}\) that maps uncalibrated predictions \(\hat{y}\) to calibrated scores \(y\).
Under GIRB, 
we fit separate isotonic calibrators within each group: $y_{k} \;=\; h_{g,k}\!\big(\hat{y}_{k}\big)$, for $k=1,\ldots,2K$,
where each \(h_{g,k}(\cdot)\) is learned on a held-out calibration set restricted to group \(g\).
During inference, we compute the embedding, obtain \(\hat{\mathbf{y}}=f_{\theta}(E_{\psi}(d,s))\), determine \(g=\phi(d,s)\), and map each dimension $k$ of $\hat{\mathbf{y}}$ using its corresponding calibrator $h_{g,k}$.

\noindent{\bf Remark.}
A global calibration approach can ignore important distinctions across different content types, as examples with similar raw confidence scores but different semantic characteristics may require different calibration adjustments.
Group-wise calibration addresses this limitation through a two-stage process: first, it clusters examples in the embedding space to identify semantically similar groups, and then learns separate calibration mappings tailored to each group's specific characteristics.
Consequently, the advantage of group-wise calibration is that it handles this local heterogeneity by first grouping the data via clustering in the embedding space and then learning a group-based mapping, rather than imposing a single global mapping.
Since calibration acts on the output of a score model, GIRB is model-agnostic and can be used with any score model \(f_{\theta}(\cdot)\); from simple linear models to more complex deep neural networks.


\noindent{\bf Training and inference.}
{\em Training:} Fit \(f_{\theta}\) by minimizing a multi-output regression loss ({\em e.g.}, MSE) on \(\hat{\mathbf{y}}\) against the \(2K\) model-based target $\cgt$ per $(d,s)$ pair; freeze \(\phi\) after clustering using the training embeddings; then fit \(h_{g,k}\) per group and per target on a held-out calibration split.
{\em Inference:} Given one \((d,s)\), compute \(E_{\psi}(d,s)\), obtain \(\hat{\mathbf{y}}=f_{\theta}(E_{\psi}(d,s))\), assign \(g=\phi(d,s)\), and output \(\mathbf{y}\) by applying \(h_{g,k}\) component-wise. This realizes both individual and average proxy scores in \emph{single pass} without multiple reference summaries.


\section{Experiments}
\noindent{\bf Datasets.} 
We evaluate the performance of our approach in two domains: summarization and question-answering (QA). 
For summarization, we use the FeedSum dataset \citep{song2024learning}, which covers documents from various domains.
The dataset consists of model generated summaries from multiple systems, 
each annotated with fine-grained FineSurE scores \citep{song2024finesure} on three dimensions: completeness, conciseness, and faithfulness. 
To cover a wider range of evaluation dimensions, we further obtain consistency, fluency, and coherence using UniEval~\citep{lee2024unisumeval}.

For QA, we follow \citet{manggala2024qa} and evaluate in six QA datasets:
SciQ \citep{welbl2017crowdsourcing}, TriviaQA \citep{joshi2017TriviaQA}, MathQA \citep{amini2019mathqa}, TruthfulQA \citep{lin2021truthfulqa}, MMLU \citep{hendrycks2020measuring}, and OpenBookQA \citep{mihaylov2018can}. 
Appendix Table~\ref{tab:dataset_summary} summarizes these datasets.

\noindent{\bf Baselines.}
In the summarization task, since our calibration targets are continuous-valued variables, calibration methods designed for discrete variables are not applicable, thus we compare to two basic settings: no calibration (None), and QA binning (QAB) \citep{manggala2024qa}.

In the QA task, we consider the following baselines: 
no recalibration (None), histogram binning (B)~\citep{gupta2021distribution}, Platt scaling (S)~\citep{platt1999probabilistic}, scaling-binning (S-B)~\citep{kumar2019verified}, QA binning (QAB), scaling QA binning (S-QAB),  hierarchical scaling QA binning (HS-QAB)~\citep{manggala2024qa}.
These baselines constitute the state-of-the-art approaches in {\em post hoc} calibration.

\noindent {\bf Evaluation Metrics.}
For the summarization task, we employ several performance metrics.

\noindent {\em Expected Calibration Error (ECE):} \(\mathrm{ECE}= \mathbb{E}_{D,S}[\,|\mathbb{E}[T\mid Y] - Y|\,]\), where $T$ and $Y$ denote ground-truth scores and proxy scores, respectively.

\noindent{\em Calibration Slope:} Given a set of ground-truth scores $t$ and proxy scores $y$, we first sort $y$ and partition it into $B$ bins according to quantiles. For each bin $b=1,\ldots,B$, we compute the average model score and the average proxy score, and then fit a linear regression on these aggregated pairs: the slope quantifies the degree of linear alignment between proxy scores and model-based scores, with a slope close to $1$ indicating good calibration and deviations from $1$ suggesting miscalibration.

\noindent {\em Group-Conditional Expected Calibration Error:} The applicability of average calibration methods such as ECE is limited, since they average over all input pairs regardless of domain; inputs can span diverse domains such as news, politics or medicine.
To address this, we also adopt group-based calibration metrics following \citet{manggala2024qa}, defining the group conditional expected calibration error (GECE) as \(\mathrm{GECE}= \mathbb{E}_{D,S}[\,|\mathbb{E}[T\mid Y,\phi(D,S)] - Y|\,]\), where $\phi(D,S)$ denotes the group assignment function learned during training. 

\begin{figure*}[t]
    \centering
    \includegraphics[width=1\textwidth]{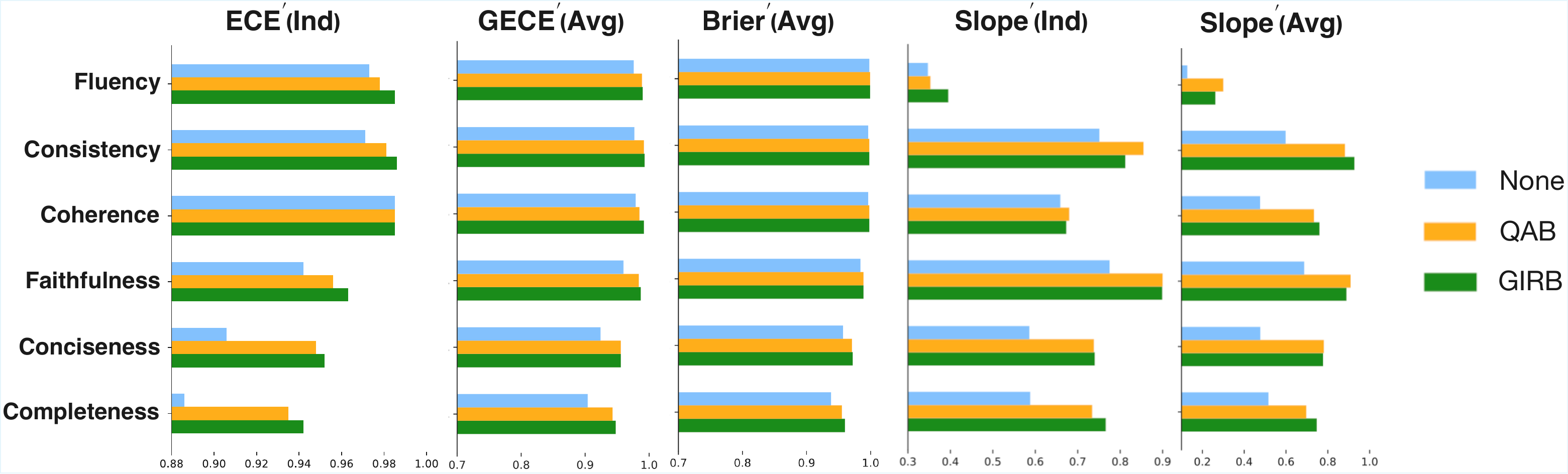}
    \vspace{-4mm}
    \caption{
    Results comparing calibration methods across dimensions and metrics. The left two plots show cases where our method achieves its best performance relative to others in terms of number of wins. The middle plot represent ties between our method and other methods. The right two plots show cases where our method performs slightly worse than its best case, yet still outperforms other methods.
    }
    \label{fig:barplot_sum_1}
\end{figure*}

\noindent{{\em Brier Score:} We also use the popular Brier score~\citep{brier1950verification}, which measures the accuracy of probabilistic predictions; although it can be decomposed into calibration and refinement components~\citep{blattenberger1985separating}, it does not directly assess calibration quality, so models with low squared error may still be poorly calibrated. 

\noindent{\em Accuracy:} Our model predicts two types of scores: an individual and an average.
We quantify whether the individual score is higher or lower than the average score, both in prediction and ground truth, thus framing it as a binary classification problem and reporting accuracy.

For QA tasks, we adopt the same calibration metrics (ECE, GECE, and Brier).
We also evaluate selective QA performance using the Area Under the Accuracy-Confidence Curve (AUAC)~\citep{el2010foundations}, where the $x$ axis corresponds to confidence thresholds $\tau \in [0,1]$, the $y$ axis is the accuracy of predictions with confidence exceeding $\tau$, and AUAC is defined as the area under this curve; a higher AUAC indicates that confidence scores better distinguish correct from incorrect answers, enabling more effective selective QA.

To make comparisons more intuitive (\emph{larger values indicate better performance}), we transform the evaluation metrics into their higher-is-better variants.
Thus, in Tables~\ref{tab:calibrator_summary},~\ref{tab:calibrator_summary_MMLU}, the summary Tables in Appendices~\ref{app:summarytabsum} and \ref{app:summarytabqa}, Figures~\ref{fig:barplot_sum_1} and \ref{fig:bar_plot_qa}, and the Figures in Appendices~\ref{app:figqa} and~\ref{app:figsum}, we report the following transformed metrics: \ECEc~$= 1 - $ ECE, \GECEc~$= 1 - $ GECE, \Brierc~$= 1 - $ Brier, \Slopec~$=1 - \lvert 1 - $ Slope $\rvert$.
\Slopec ~attains \(1\) when Slope\(~=1\) and decreases symmetrically as Slope departs from \(1\).
For each evaluation metric, we use (Ind) to denote the performance based on individual scores and (Avg) to denote the performance based on average scores.


\begin{table}[t]
\scalebox{0.7}{
\begin{tabular}{lcccccc}
\hline
Method & \multicolumn{2}{c}{None} & \multicolumn{2}{c}{QAB}                & \multicolumn{2}{c}{GIRB} \\
Metric       & Gain & Wins & Gain & Wins           & Gain & Wins                                          \\ \hline
\ECEc (Ind)         & -      & 1             & +0.022 & 1                           & \textbf{+0.027}                           & \textbf{6}                  \\
\GECEc (Ind)     & -      & 0             & +0.028 & 2                           & \textbf{+0.030}                           & \textbf{5}                  \\
\Brierc (Ind)       & -      & 1             & +0.005 & 2                           & \textbf{+0.008}                           & \textbf{5}                  \\
\Slopec (Ind)       & -      & 0             & +0.147 & \textbf{3} & \textbf{+0.161}                           & \textbf{3}                  \\
Accuracy (Ind)     & -      & \textbf{3}      & -0.043 & 0                           & \textbf{+0.017}                           & \textbf{3}                  \\
\ECEc (Avg)     & -      & 0             & +0.033 & 4 & \textbf{+0.035} & \textbf{6}                           \\
\GECEc (Avg) & -      & 0             & +0.023 & 1                           & \textbf{+0.026}                           & \textbf{6}                  \\
\Brierc (Avg)   & -      & 0             & +0.007 & 4 & \textbf{+0.008} & \textbf{6}                           \\
\Slopec (Avg)  & -      & 0             & \textbf{+0.612} & \textbf{3}                  & +0.594                           & \textbf{3} \\ 
\cdashline{1-7}
Mean      & -      & 0.56          & +0.093 & 2.22                        & \textbf{+0.101}                                  & \textbf{4.78}                        \\ \hline

\end{tabular}}
\caption{Summary of performance across calibration methods for each metric. For each metric, we report two indicators. \textbf{Gain} is the average improvement compared to ``None'', and \textbf{Wins} is the number of dimensions where a given method outperforms the others. In cases of tied first place, all top-ranking methods are counted as wins.}
\label{tab:calibrator_summary}
\vspace{-6mm}
\end{table}

\noindent {\bf Training.}
For the summarization task dataset, we split the dataset into four subsets with a ratio of 30:30:30:10.
The first subset is used to construct the clustering algorithm (KMeans), the second subset is used to train a linear regression model to predict raw scores, the third subset is reserved for post-hoc calibration training, and the fourth subset is used for testing.  
For each document-summary pair, we first obtain embedding representations using the Qwen3-8B embedding model~\citep{qwen3embedding}, following the \texttt{Instruct-DS-Prompt} format described in Appendix~\ref{appendix:prompt_emb}.
These embeddings are then fed into a multi-output linear regression layer to predict $2K$ dimensional model-based scores.
We observed empirically that a simple linear layer performed similarly to more complex layers.
The regression model is trained with batch size $500$, learning rate $1 \times 10^{-3}$, and for $50$ epochs.  

For the QA task, in order to compare with the other baseline results, we strictly followed the settings of the environment and the hyper-parameters in \citet{manggala2024qa}.
Following \citet{manggala2024qa}, we use KD-Tree for the clustering in GIRB.
Note that unlike summarization task, we do not need to use the score model in QA task, and instead we use LLMs (Gemma and Mistral) with different prompts (\texttt{ling1s-topk} and \texttt{verb1s-topk}) to produce the scores for assessing the confidence of a given QA pair.
We provide more details about other hyperparameters in Appendix~\ref{app:hyperparams}.

\begin{figure*}[t!]
    \centering
    \includegraphics[width=1\textwidth]{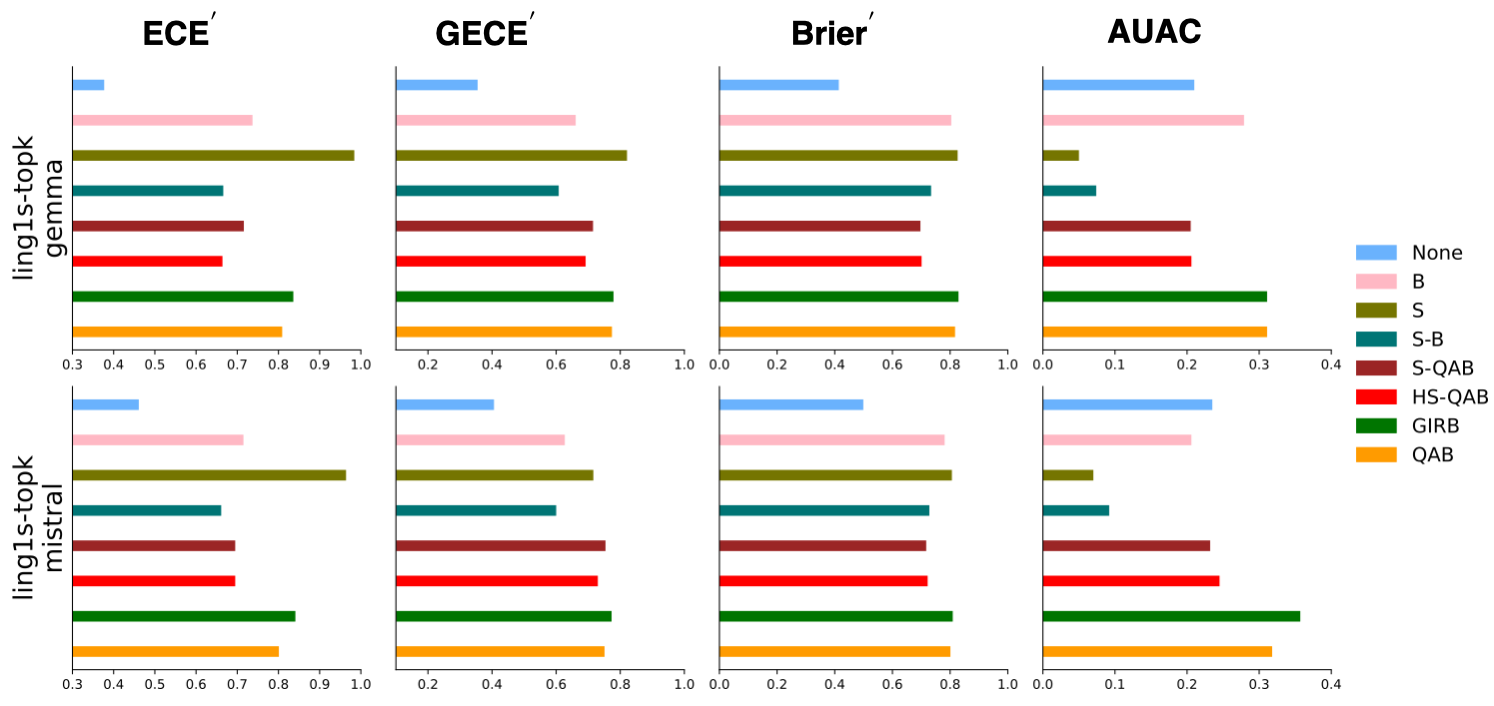}
    \vspace{-6mm}
    \caption{
    Results comparing calibration methods over all metrics in the MMLU dataset.
    }
    \label{fig:bar_plot_qa}
\end{figure*}

\noindent{\bf Summarization Results.} 
%
Figure~\ref{fig:barplot_sum_1} illustrates the performance of different calibration methods over multiple evaluation metrics and dimensions.
These visualizations allow direct comparison of each method under various settings.

To quantitatively compare the relative strengths of each method, we compute two indicators for each metric: the number of dimensions where a given method outperforms the others (Wins), and the corresponding relative improvement (Gain).
In the summary Table~\ref{tab:calibrator_summary}, ``Wins'' indicates the number of dimensions where a method achieves the best performance.
For example, for the \ECEc~metric, if GIRB achieves the top performance in four dimensions, the corresponding cell shows ``4''.
The main value in each cell represents the average relative improvement over all dimensions, calculated as $\frac{{y}_k - \hat{y}_k}{\hat{y}_k}$, where $\hat{y}_k$ is the uncalibrated score in dimension $k$, and ${y}_k$ is the calibrated score. 
Positive values indicate improvements over the baseline, while negative values mean performance drops, summarizing each method’s overall effectiveness.

In Figure~\ref{fig:barplot_sum_1}, we show the two best performing metrics on the left, the two relatively weaker metrics on the right, and a middle metric between them.
The ordering of good versus poor performance is determined by the ``Wins'' in Table~\ref{tab:calibrator_summary}.
As shown, QAB and GIRB outperform the uncalibrated baseline, consistently achieving higher scores on the conciseness and completeness dimensions.

In Table~\ref{tab:calibrator_summary}, GIRB achieves the highest number of wins in almost all metrics, except \Slopec (Ind), \Slopec (Avg), and Accuracy, where it ties with other methods.
On average, GIRB wins 4.78 times, more than double that of the second-best method, QAB, which achieves 2.22.

By combining the bar plots with the summary table, we provide both qualitative and quantitative perspectives: the figures visualize dimension-level performance for each metric, while the table summarizes wins and average improvement.
These complementary analyses offer a comprehensive understanding of the effectiveness of each calibration method in different evaluation criteria. 
Detailed results for all settings are provided in Appendix~\ref{app:sum}.

\begin{table*}[t]
\centering
\scalebox{0.69}{
\begin{tabular}{lcccccccccccccccc}
\hline
& \multicolumn{2}{c}{None} & \multicolumn{2}{c}{S-B} & \multicolumn{2}{c}{S} & \multicolumn{2}{c}{B} & \multicolumn{2}{c}{QAB} & \multicolumn{2}{c}{HS-QAB} & \multicolumn{2}{c}{S-QAB} & \multicolumn{2}{c}{GIRB} \\
Metric & Gain & Wins & Gain & Wins & Gain & Wins & Gain & Wins & Gain & Wins & Gain & Wins & Gain & Wins & Gain & Wins \\ \hline
\ECEc     & 0 & 0 & +0.695 & 0 & \textbf{+1.424} & \textbf{4} & +0.864 & 0 & +1.061 & 0 & +0.739 & 0 & +0.780 & 0 & +1.145 & 0 \\
\GECEc & 0 & 0 & +0.654 & 0 & +1.077 & 1 & +0.814 & 0 & +1.130 & 0 & +0.968 & 0 & +1.030 & 0 & \textbf{+1.167} & \textbf{3} \\
\Brierc   & 0 & 0 & +0.700 & 0 & +0.889 & 0 & +0.845 & 0 & +0.873 & 0 & +0.648 & 0 & +0.642 & 0 & \textbf{+0.901} & \textbf{4} \\
AUAC    & 0 & 0 & -0.606 & 0 & -0.703 & 0 & +0.270 & 0 & +0.429 & 1 & +0.039 & 0 & +0.037 & 0 & \textbf{+0.531} & \textbf{3} \\ 
\cdashline{1-17}
Mean & 0 & 0 & +0.361 & 0 & +0.672 & 0.2 & +0.698 & 0 & +0.873 & 0.2 & +0.599 & 0 & +0.622 & 0 & \textbf{+0.936} & \textbf{2.5} \\ \hline
\end{tabular}}
\caption{Summary of performance across calibration methods for each metric in MMLU dataset. For each metric, we report two indicators: \textbf{Gain} is the average improvement compared to ``None'', and \textbf{Wins} is the number of settings ({\em e.g.}, ling1s-topk) where a given method outperforms the others. In cases of tied first place, all top-ranking methods are counted as wins.}
\label{tab:calibrator_summary_MMLU}
\vspace{-4mm}
\end{table*}

\noindent{\bf Ablation Study.}
We denote a setting as a combination of embedding model, prompting strategy and clustering method.
In the main results, we use Qwen, Prompt and KMeans, respectively.
We perform an ablation study by varying one component at a time while keeping the others fixed to assess its impact.
We conducted experiments with different clustering methods, prompting strategies and embedding models, as summarized in Appendix Tables~\ref{tab:calibrator_summary_kdtree}, \ref{tab:calibrator_summary_Pair}, and \ref{tab:calibrator_summary_emb}, respectively.

From these results, we observe that:
$i$) Over all alternative settings, our method consistently outperforms the others in terms of the average number of wins.
$ii$) Although the average Gain is not a strictly meaningful metric, as it averages values on different scales, our method (GIRB) still achieves the best overall performance, except in the KD-Tree setting.
$iii$) When examining the contribution of individual components, the average number of wins follows the ranking: KMeans > KD-Tree, Instruct-DS-Prompt > DS-Prompt, and Qwen3-8B > DistilBERT > Gemma-2B, suggesting that our default configuration represents the optimal combination. 
Importantly, DistilBERT is not a large generative LLM but a lightweight embedding-based model. Its competitive performance indicates that our method does not strictly depend on large-scale LLMs, providing empirical evidence on the degree of dependence on embedding model size and demonstrating that the proposed approach remains effective even with substantially smaller models.

\noindent{\bf QA Results.}
We further validate our approach in a QA task, following the setup described in \citet{manggala2024qa}.
Based on the questions and reference answers in the dataset, we use LLMs (Gemma and Mistral) to generate answers and associated confidence scores using specific prompts (\texttt{ling1s-topk} and \texttt{verb1s-topk}).
Complete prompt specifications are provided in Appendix~\ref{appendix:prompt_qa}. 
Subsequently, we employ another large language model, LLaMA, to compare the generated answers with the reference answers and determine their correctness, 0 or 1.

This study compares the effectiveness of different calibration methods in aligning confidence scores with answer correctness.
To better illustrate the performance of each calibration method over multiple evaluation metrics and settings, we present bar plots that allow for direct visual comparison.
Figure~\ref{fig:bar_plot_qa} shows results for two setting, where answers and confidence scores were generated by Mistral and Gemma using \texttt{ling1s-topk} prompt. The complete set of results is provided in Appendix Figure~\ref{fig:bar_plot_qa_MMLU}.
As shown, GIRB achieves the best performance in all metrics except ECE, where scaling binning (S-B) performs the best, and our method ranks second.


Following the same approach as in Table~\ref{tab:calibrator_summary}, Table~\ref{tab:calibrator_summary_MMLU} summarizes ``Wins'' and ``Gains'' indicating the number of settings in which a method achieves the best performance, along with the corresponding improvement magnitude.
Considering the average number of wins across metrics, GIRB achieves 2.5, much higher than the second-best at 0.2.
In terms of average gain, GIRB obtains 0.936, also surpassing the second-best 0.873. 
We provide detailed results for various settings in Appendix~\ref{app:qa}.

\begin{table}[t]
\centering
\scalebox{0.8}{
\begin{tabular}{llllll}
\hline
 Calibrator & \ECEc   & \GECEc  & \Brierc  \\ \hline
 None       & 0.983 & 0.983 & 0.948  \\
 QAB        & 0.990 & 0.990 & \bf{0.953}  \\
 GIRB       & \bf{0.993} & \bf{0.993} & 0.940  \\ \hline
\end{tabular}}
\caption{Performance of different calibration methods in human evaluation dataset.}
\label{tab:human_study_table}
\vspace{-5mm}
\end{table}

\noindent{\bf Human Study.}
To further evaluate the effectiveness of the proposed calibration method, we leverage datasets with human annotations to examine whether the calibrated scores better align with human judgments. Specifically, we adopt the UniSumEval benchmark \citep{lee2024unisumeval}, which provides document-summary pairs annotated with both human evaluation scores and model-based scores. In our experiments, we use human faithfulness scores as ground truth score and calibrate model-based G-Eval+ scores using both QAB and GIRB. As shown in Table~\ref{tab:human_study_table}, calibrated scores provide a better proxy for human judgments, achieving higher \ECEc, \GECEc, and \Brierc scores than without calibration. Moreover, GIRB consistently outperforms QAB in terms of ECE and GECE. Overall, these results demonstrate that our calibration method yields stronger agreement with human ratings and can be effectively generalized to calibrate the model-based scores using human annotations as the ground truth.

\vspace{-0.1cm}
\section{Discussion}
\vspace{-0.1cm}


The primary contribution of this work is the introduction of the GIRB framework, which addresses the persistent issue of miscalibration in model-based evaluation metrics. Our results show that GIRB enables reliable estimation of both individual-level summary quality scores and average-level summary quality, without requiring expensive reference summaries or human annotations. This capability has important practical implications for the deployment of LLMs in real-world settings, where quality assessment must be performed efficiently and at scale. In particular, GIRB allows systems to determine whether generated outputs meet predefined quality thresholds in a single, cost-effective pass.

Another finding is that calibration improves when semantic structure is incorporated through grouping. By clustering the embedding space, GIRB accounts for differences across content types ({\em e.g.}, news versus medical text). This approach breaks a complex global calibration problem into simpler local ones, allowing the model to better capture group-specific biases and improve alignment between proxy scores and ground-truth metrics.

Furthermore, isotonic regression offers a simple and effective alternative to prior binning-based methods such as QAB. Instead of relying on fixed-width bins and manual tuning, it learns a continuous and monotonic mapping from proxy scores to calibrated values. This leads to a smoother calibration function and better performance in continuous settings such as summarization. The consistent gains across datasets suggest that GIRB is both effective and robust, making it a practical solution for calibration in model-based evaluation.

\section*{Limitations}
Despite these promising results, GIRB has several limitations. First, its performance depends on the choice of the clustering algorithm, as the quality of group assignments directly affects calibration. Furthermore, the effectiveness of the grouping stage is inherently tied to the representational power of the embedding model. If the embeddings fail to capture the semantic nuances required to distinguish between different bias distributions, the resulting calibration improvements may be limited.
Second, the current calibration procedure relies on a sufficient amount of post-calibration data, which may not always be available in practice. This limitation highlights the need for more data-efficient calibration methods. Future work may also explore more adaptive or learned grouping strategies, as well as extend the proposed approach to other modalities and evaluation settings.



\bibliography{custom}
\newpage
\appendix

\section{Model Hyperparameters}\label{app:hyperparams}
For calibration analysis, we used quantile binning with 10 equal-frequency bins to compute ECE and Slope. In the summarization task, we employed KD-Tree with a maximum depth of 8 and KMeans with 256 clusters. The large language embedding models used in this task were Qwen-3-8B and Gemma-2B. In the QA task, we conducted a hyperparameter search over KD-Tree maximum depth values ranging from 0 to 10, using Mistral-7B and Gemma-7B models to produce score, and using LLaMA2-7B to produce the ground truth correctness label.

\section{Dataset Details}
Table~\ref{tab:dataset_summary} provides the details about the summarization and QA datasets.
\begin{table}[h]
\caption{Dataset summary.}
\hspace{1mm}
\centering
\begin{tabular}{lll}
\hline
Dataset    & Size  & Type               \\ \hline
FeedSum    & 76945 & Summarization      \\
SciQ       & 11609 & QA                 \\
TriviaQA   & 11313 & QA                 \\
MathQA     & 29837 & QA                 \\
TruthfulQA & 790   & QA                 \\ 
MMLU       & 20320 & Multiple-choice QA \\
OpenBookQA & 13869 & Multiple-choice QA \\ \hline
\end{tabular}
\label{tab:dataset_summary}
\end{table}

\section{Summarization Task}\label{app:sum}

\subsection{Prompt for Obtaining Embedding in Document Summarization Task}\label{appendix:prompt_emb}

We adopt two approaches to obtain embeddings from the document and its summary. The first approach directly concatenates the document and summary pair, which we refer to as \texttt{DS-Prompt}. The second approach simulates prompting a large language model, which we refer to as \texttt{Instruct-DS-Prompt}.

\paragraph{\texttt{DS-Prompt}}
{\small \textit{
\\Document: \textcolor{blue}{<document>}\\
Summary: \textcolor{blue}{<summary>}
}}

\paragraph{\texttt{Instruct-DS-Prompt}}
{\small \textit{
\\Given a document and its summary, compute a score that estimates the summary’s quality.\\
Document: \textcolor{blue}{<document>}\\
Summary: \textcolor{blue}{<summary>}
}}

\subsection{Summary Tables}\label{app:summarytabsum}
This section presents the additional summary table results for the summarization task.
For each metric, we report ``Gain'' (average improvement over the uncalibrated baseline) and ``Wins'' (number of dimensions where the method outperforms others), enabling comparison of calibration effectiveness across prompting strategies.
Table~\ref{tab:calibrator_summary_kdtree} presents a summary of performance across calibration methods for each metric using KMeans and KD-Tree. 
Table~\ref{tab:calibrator_summary_Pair} summarizes calibration method performance using \texttt{DS-Prompt} and \texttt{Instruct-DS-Prompt} settings. 
Table~\ref{tab:calibrator_summary_emb} reports the summary of performance across calibration methods using different embedding model settings.

\begin{table*}[h]
\centering
\caption{Summary of performance across calibration methods for each metric. For each metric, we report two indicators: i. Gain: average improvement compared to ``None'', and ii. Wins: the number of dimensions where a given method outperforms the others using KMean and KD-Tree setting.}
\scalebox{0.7}{\label{tab:calibrator_summary_kdtree}
\begin{tabular}{lcccccclcccccc}
\hline
\multicolumn{7}{l}{KMean}                                                                             & \multicolumn{7}{l}{KD-Tree}                                                                            \\ \cdashline{1-14}
Method       & \multicolumn{2}{c}{None} & \multicolumn{2}{c}{QAB}      & \multicolumn{2}{c}{GIRB}     & Method       & \multicolumn{2}{c}{None} & \multicolumn{2}{c}{QAB}      & \multicolumn{2}{c}{GIRB}     \\
Metric       & Gain     & Wins          & Gain            & Wins       & Gain            & Wins       & Metric       & Gain        & Wins       & Gain            & Wins       & Gain            & Wins       \\ \cdashline{1-14}
\ECEc (Ind)          & -        & 1             & +0.022          & 1          & \textbf{+0.027} & \textbf{6} & \ECEc (Ind)          & -           & 1          & +0.022          & 3          & \textbf{+0.023} & \textbf{4} \\
\GECEc (Ind)      & -        & 0             & +0.028          & 2          & \textbf{+0.030} & \textbf{5} & \GECEc (Ind)      & -           & 0          & +0.026          & 0          & \textbf{+0.030} & \textbf{6} \\
\Brierc (Ind)        & -        & 1             & +0.005          & 2          & \textbf{+0.008} & \textbf{5} & \Brierc (Ind)        & -           & 2          & +0.001          & 3          & \textbf{+0.002} & \textbf{4} \\
\Slopec (Ind)        & -        & 0             & +0.147          & \textbf{3} & \textbf{+0.161} & \textbf{3} & \Slopec (Ind)        & -           & 1          & \textbf{+0.255} & \textbf{4} & +0.126          & 1          \\
Accuracy     & -        & \textbf{3}    & -0.043          & 0          & \textbf{+0.017} & \textbf{3} & Accuracy     & -           & 1          & +0.030          & 0          & \textbf{+0.067} & \textbf{5} \\
\ECEc (Avg)     & -        & 0             & +0.033          & 4          & \textbf{+0.035} & \textbf{6} & \ECEc (Avg)     & -           & 0          & +0.032          & 1          & \textbf{+0.036} & \textbf{5} \\
\GECEc (Avg) & -        & 0             & +0.023          & 1          & \textbf{+0.026} & \textbf{6} & \GECEc (Avg) & -           & 0          & +0.029          & 2          & \textbf{+0.031} & \textbf{5} \\
\Brierc (Avg)   & -        & 0             & +0.007          & 4          & \textbf{+0.008} & \textbf{6} & \Brierc (Avg)   & -           & 0          & +0.006          & 3          & \textbf{+0.007} & \textbf{6} \\
\Slopec (Avg)   & -        & 0             & \textbf{+0.612} & \textbf{3} & +0.594          & \textbf{3} & \Slopec (Avg)   & -           & 0          & \textbf{+1.075} & \textbf{6} & +0.899          & 0          \\ \cdashline{1-14}
Mean      & -        & 0.56          & +0.093          & 2.22       & \textbf{+0.101}          & \textbf{4.78}       & Mean      & -           & 0.56       & \textbf{+0.164}          & 2.44       & +0.136          & \textbf{4}          \\ \hline
\end{tabular}}
\end{table*}

\begin{table*}[h]
\centering
\caption{Summary of performance across calibration methods for each metric using either \texttt{DS-Prompt} or \texttt{Instruct-DS-Prompt} setting. For each metric, we report two indicators: i. Gain: average improvement compared to ``None'', and ii. Wins: the number of dimensions where a given method outperforms the others.}
\scalebox{0.7}{\label{tab:calibrator_summary_Pair}
\begin{tabular}{lcccccclcccccc}
\hline
\multicolumn{7}{l}{Instruct-DS-Prompt}                                                                      & \multicolumn{7}{l}{DS-Prompt}                                                                              \\ \cdashline{1-14}
Method       & \multicolumn{2}{c}{None} & \multicolumn{2}{c}{QAB}      & \multicolumn{2}{c}{GIRB}     & Method       & \multicolumn{2}{c}{None} & \multicolumn{2}{c}{QAB}      & \multicolumn{2}{c}{GIRB}     \\
Metric       & Gain     & Wins          & Gain            & Wins       & Gain            & Wins       & Metric       & Gain     & Wins          & Gain            & Wins       & Gain            & Wins       \\ \cdashline{1-14}
\ECEc (Ind)          & -        & 1             & +0.022          & 1          & \textbf{+0.027} & \textbf{6} & \ECEc (Ind)          & -        & 0             & \textbf{+0.039} & \textbf{4} & \textbf{+0.035} & \textbf{3} \\
\GECEc (Ind)      & -        & 0             & +0.028          & 2          & \textbf{+0.030} & \textbf{5} & \GECEc (Ind)      & -        & 0             & \textbf{+0.043} & \textbf{3} & \textbf{+0.042} & \textbf{3} \\
\Brierc (Ind)        & -        & 1             & +0.005          & 2          & \textbf{+0.008} & \textbf{5} & \Brierc (Ind)        & -        & 1             & +0.010          & 1          & \textbf{+0.011} & \textbf{4} \\
\Slopec (Ind)        & -        & 0             & +0.147          & \textbf{3} & \textbf{+0.161} & \textbf{3} & \Slopec (Ind)        & -        & 0             & \textbf{+0.556} & \textbf{5} & +0.443          & 1          \\
Accuracy     & -        & \textbf{3}    & -0.043          & 0          & \textbf{+0.017} & \textbf{3} & Accuracy     & -        & \textbf{3}    & -0.072          & 0          & \textbf{+0.035} & \textbf{3} \\
\ECEc (Avg)     & -        & 0             & +0.033          & 4          & \textbf{+0.035} & \textbf{6} & \ECEc (Avg)     & -        & 0             & +0.052          & 0          & \textbf{+0.057} & \textbf{6} \\
\GECEc (Avg) & -        & 0             & +0.023          & 1          & \textbf{+0.026} & \textbf{6} & \GECEc (Avg) & -        & 0             & +0.041          & 0          & \textbf{+0.046} & \textbf{6} \\
\Brierc (Avg)   & -        & 0             & +0.007          & 4          & \textbf{+0.008} & \textbf{6} & \Brierc (Avg)   & -        & 0             & +0.015          & 4          & \textbf{+0.016} & \textbf{6} \\
\Slopec (Avg)   & -        & 0             & \textbf{+0.612} & \textbf{3} & +0.594          & \textbf{3} & \Slopec (Avg)   & -        & 0             & \textbf{-5.673} & \textbf{5} & \textbf{-4.431} & 1          \\ \cdashline{1-14}
Mean      & -        & 0.56          & +0.093          & 2.22       & \textbf{+0.101}          & \textbf{4.78}       & Mean      & -        & 0.44          & -0.554          & 2.44       & \textbf{-0.416}          & \textbf{3.67}       \\ \hline
\end{tabular}}
\end{table*}

\begin{table*}[h]
\centering
\caption{Summary of performance across calibration methods for each metric. For each metric, we report two indicators: i. Gain: average improvement compared to ``None'', and ii. Wins: the number of dimensions where a given method outperforms the others using different embedding model setting.}
\scalebox{0.5}{\label{tab:calibrator_summary_emb}
\begin{tabular}{lcccccclcccccclcccccc}
\hline
\multicolumn{7}{l}{Qwen-8B}                                                                           & \multicolumn{7}{l}{DistilBert}                                                                        & \multicolumn{7}{l}{Gemma-2B}                                                                          \\ \cdashline{1-21}
Method       & \multicolumn{2}{c}{None} & \multicolumn{2}{c}{QAB}      & \multicolumn{2}{c}{GIRB}     & Method       & \multicolumn{2}{c}{None} & \multicolumn{2}{c}{QAB}      & \multicolumn{2}{c}{GIRB}     & Method       & \multicolumn{2}{c}{None} & \multicolumn{2}{c}{QAB}      & \multicolumn{2}{c}{GIRB}     \\
Metric       & Gain     & Wins          & Gain            & Wins       & Gain            & Wins       & Metric       & Gain     & Wins          & Gain            & Wins       & Gain            & Wins       & Metric       & Gain     & Wins          & Gain            & Wins       & Gain            & Wins       \\ \cdashline{1-21}
\ECEc (Ind)          & -        & 1             & +0.022          & 1          & \textbf{+0.027} & \textbf{6} & \ECEc (Ind)          & -        & 0             & \textbf{+0.028} & \textbf{2} & \textbf{+0.029} & \textbf{5} & \ECEc (Ind)          & -        & 1             & +0.015          & 2          & \textbf{+0.018} & \textbf{4} \\
\GECEc (Ind)      & -        & 0             & +0.028          & 2          & \textbf{+0.030} & \textbf{5} & \GECEc (Ind)      & -        & 0             & \textbf{+0.029} & \textbf{1} & \textbf{+0.032} & \textbf{5} & \GECEc (Ind)      & -        & 0             & +0.023          & 2          & \textbf{+0.026} & \textbf{6} \\
\Brierc (Ind)        & -        & 1             & +0.005          & 2          & \textbf{+0.008} & \textbf{5} & \Brierc (Ind)        & -        & 1             & +0.006          & \textbf{4} & \textbf{+0.008} & \textbf{4} & \Brierc (Ind)        & -        & 2             & \textbf{+0.009} & \textbf{3} & \textbf{+0.009} & 2          \\
\Slopec (Ind)        & -        & 0             & +0.147          & \textbf{3} & \textbf{+0.161} & \textbf{3} & \Slopec (Ind)        & -        & 1             & \textbf{+0.316} & \textbf{3} & +0.289          & 2          & \Slopec (Ind)        & -        & 1             & +0.077          & \textbf{4} & \textbf{+0.089} & 1          \\
Accuracy     & -        & \textbf{3}    & -0.043          & 0          & \textbf{+0.017} & \textbf{3} & Accuracy     & -        & \textbf{1}    & -0.043          & 0          & \textbf{+0.033} & \textbf{5} & Accuracy     & -        & \textbf{4}    & -0.096          & 0          & \textbf{-0.013} & 2          \\
\ECEc (Avg)     & -        & 0             & +0.033          & 4          & \textbf{+0.035} & \textbf{6} & \ECEc (Avg)     & -        & 0             & +0.041          & 3          & \textbf{+0.042} & \textbf{5} & \ECEc (Avg)     & -        & 0             & +0.039          & 1          & \textbf{+0.042} & \textbf{6} \\
\GECEc (Avg) & -        & 0             & +0.023          & 1          & \textbf{+0.026} & \textbf{6} & \GECEc (Avg) & -        & 0             & +0.033          & 1          & \textbf{+0.036} & \textbf{6} & \GECEc (Avg) & -        & 0             & +0.028          & 1          & \textbf{+0.031} & \textbf{6} \\
\Brierc (Avg)   & -        & 0             & +0.007          & 4          & \textbf{+0.008} & \textbf{6} & \Brierc (Avg)   & -        & 0             & \textbf{+0.013} & 3          & \textbf{+0.013} & \textbf{5} & \Brierc (Avg)   & -        & 1             & +0.012          & 3          & \textbf{+0.013} & \textbf{6} \\
\Slopec (Avg)   & -        & 0             & \textbf{+0.612} & \textbf{3} & +0.594          & \textbf{3} & \Slopec (Avg)   & -        & 0             & \textbf{+0.905} & \textbf{4} & \textbf{+0.887} & 2          & \Slopec (Avg)   & -        & 0             & +0.731          & \textbf{3} & \textbf{+0.738} & \textbf{3} \\ \cdashline{1-21}
Mean      & -        & 0.56          & +0.093          & 2.22       & \textbf{+0.101}          & \textbf{4.78}       & Mean      & -        & 0.33          & +0.148          & 2.33       & \textbf{+0.152}          & \textbf{4.33}       & Mean      & -        & 1             & +0.093          & 2.11       & \textbf{+0.106}          & \textbf{4}          \\ \hline
\end{tabular}}
\end{table*}

\subsection{Complete Results Tables}\label{app:fulltabsum}
In this section, we present the complete set of results of all the experiments for the summarization task.
We provide a comprehensive performance comparison of multiple calibration methods across various settings using the FeedSum dataset for the summarization task in Table~\ref{tab:summary_main_table}.
The performance comparison of different clustering methods and calibration methods across various dimensions is provided in Table~\ref{tab:summary_ablation_table_clustering}.
We compare the performance of different embedding models and calibration methods across various dimensions in Table~\ref{tab:summary_ablation_table_embedding}.
Table~\ref{tab:summary_ablation_table_strategy} shows the performance comparison of different prompting strategies and calibration methods across various dimensions.

\begin{table*}[h]
\centering
\caption{Performance of different calibration methods across various setting in FeedSum dataset.
}
\scalebox{0.8}{\label{tab:summary_main_table}
\begin{tabular}{lllllll}
\hline
Dimension         & Calibrator & ECE                        & GECE                    & Brier                      & Slope                      & Accuracy                   \\ \hline
Completeness (Ind) & None       & 0.114 $\pm$ 0.034          & 0.125 $\pm$ 0.24           & 0.106 $\pm$ 0.040          & 0.588 $\pm$ 0.276          & 0.631 $\pm$ 0.122          \\
Completeness (Ind) & QAB        & 0.065 $\pm$ 0.026          & 0.084 $\pm$ 0.022          & 0.090 $\pm$ 0.017          & 0.734 $\pm$ 0.188          & 0.668 $\pm$ 0.108          \\
Completeness (Ind) & GIRB       & \textbf{0.058 $\pm$ 0.012} & \textbf{0.082 $\pm$ 0.031} & \textbf{0.084 $\pm$ 0.020} & \textbf{0.766 $\pm$ 0.311} & \textbf{0.702 $\pm$ 0.139} \\ \hline
Conciseness (Ind) & None       & 0.094 $\pm$ 0.025          & 0.11 $\pm$ 0.023           & 0.098 $\pm$ 0.038          & 0.586 $\pm$ 0.274          & \textbf{0.742 $\pm$ 0.145} \\
Conciseness (Ind) & QAB        & 0.052 $\pm$ 0.024          & \textbf{0.077 $\pm$ 0.021} & 0.088 $\pm$ 0.018          & 0.738 $\pm$ 0.265          & 0.668 $\pm$ 0.118          \\
Conciseness (Ind) & GIRB       & \textbf{0.048 $\pm$ 0.023} & 0.080 $\pm$ 0.017          & \textbf{0.084 $\pm$ 0.020} & \textbf{0.740 $\pm$ 0.150} & 0.731 $\pm$ 0.119          \\ \hline
Faithfulness (Ind) & None       & 0.058 $\pm$ 0.020          & 0.098 $\pm$ 0.25           & 0.085 $\pm$ 0.017          & 0.775 $\pm$ 0.267          & 0.790 $\pm$ 0.154          \\
\hline
\end{tabular}}
\end{table*}

\begin{table*}[h]
\centering
\caption{The performance of different clustering methods and calibration methods across various dimensions.
}
\scalebox{0.65}{\label{tab:summary_ablation_table_clustering}
\begin{tabular}{llllllll}
\hline
Cluster\_method & Dimension         & Calibrator & ECE                        & GECE                    & Brier                      & Slope                      & Accuracy                   \\ \hline
KD-Tree          & Completeness (Ind) & None       & 0.114 $\pm$ 0.034          & 0.112 $\pm$ 0.014          & 0.106 $\pm$ 0.044          & 0.588 $\pm$ 0.243          & 0.631 $\pm$ 0.105          \\
KD-Tree          & Completeness (Ind) & QAB        & \textbf{0.058 $\pm$ 0.025} & 0.071 $\pm$ 0.013          & 0.097 $\pm$ 0.032          & \textbf{0.730 $\pm$ 0.169} & 0.662 $\pm$ 0.146          \\
KD-Tree          & Completeness (Ind) & GIRB       & \textbf{0.058 $\pm$ 0.012} & \textbf{0.070 $\pm$ 0.024} & \textbf{0.094 $\pm$ 0.025} & 0.695 $\pm$ 0.288          & \textbf{0.691 $\pm$ 0.131} \\ \hline
KD-Tree          & Conciseness (Ind) & None       & 0.094 $\pm$ 0.025          & 0.117 $\pm$ 0.020          & 0.098 $\pm$ 0.033          & 0.586 $\pm$ 0.290          & 0.642 $\pm$ 0.107          \\
KD-Tree          & Conciseness (Ind) & QAB        & 0.056 $\pm$ 0.025          & 0.073 $\pm$ 0.014          & 0.096 $\pm$ 0.040          & 0.708 $\pm$ 0.196          & 0.688 $\pm$ 0.124          \\
KD-Tree          & Conciseness (Ind) & GIRB       & \textbf{0.055 $\pm$ 0.023} & \textbf{0.067 $\pm$ 0.024} & \textbf{0.091 $\pm$ 0.021} & \textbf{0.716 $\pm$ 0.193} & \textbf{0.694 $\pm$ 0.186} \\ \hline
KD-Tree          & Faithfulness (Ind) & None       & 0.058 $\pm$ 0.020          & 0.085 $\pm$ 0.017          & 0.085 $\pm$ 0.028          & 0.775 $\pm$ 0.344          & 0.690 $\pm$ 0.110          \\
\hline
\end{tabular}}
\end{table*}

\begin{table*}[h]
\centering
\caption{The performance of different embedding models and calibration methods across various dimensions.
}
\scalebox{0.45}{\label{tab:summary_ablation_table_embedding}
\begin{tabular}{llllllll}
\hline
Embedding\_model & Dimension         & Calibrator & ECE                        & GECE                    & Brier                      & Slope                      & Accuracy                   \\ \hline
DistilBERT       & Completeness (Ind) & None       & 0.116 $\pm$ 0.034          & 0.121 $\pm$ 0.018          & 0.112 $\pm$ 0.029          & 0.598 $\pm$ 0.122          & 0.629 $\pm$ 0.129          \\
DistilBERT       & Completeness (Ind) & QAB        & 0.059 $\pm$ 0.025          & \textbf{0.062 $\pm$ 0.024} & \textbf{0.091 $\pm$ 0.026} & 0.754 $\pm$ 0.233          & 0.607 $\pm$ 0.111          \\
DistilBERT       & Completeness (Ind) & GIRB       & \textbf{0.054 $\pm$ 0.013} & 0.067 $\pm$ 0.030          & \textbf{0.091 $\pm$ 0.032} & \textbf{0.809 $\pm$ 0.222} & \textbf{0.641 $\pm$ 0.100} \\ \hline
DistilBERT       & Conciseness (Ind) & None       & 0.118 $\pm$ 0.026          & 0.104 $\pm$ 0.026          & 0.108 $\pm$ 0.044          & 0.550 $\pm$ 0.173          & \textbf{0.696 $\pm$ 0.111} \\
DistilBERT       & Conciseness (Ind) & QAB        & 0.070 $\pm$ 0.027          & 0.076 $\pm$ 0.027          & 0.102 $\pm$ 0.016          & 0.562 $\pm$ 0.204          & 0.622 $\pm$ 0.163          \\
DistilBERT       & Conciseness (Ind) & GIRB       & \textbf{0.069 $\pm$ 0.024} & \textbf{0.075 $\pm$ 0.016} & \textbf{0.091 $\pm$ 0.012} & \textbf{0.663 $\pm$ 0.105} & 0.669 $\pm$ 0.106          \\ \hline
DistilBERT       & Faithfulness (Ind) & None       & 0.088 $\pm$ 0.027          & 0.11 $\pm$ 0.016           & 0.112 $\pm$ 0.024          & 0.534 $\pm$ 0.096          & 0.656 $\pm$ 0.118          \\
\hline
\end{tabular}}
\end{table*}

\begin{table*}[h]
\centering
\caption{The performance of different prompting strategy and calibration methods across various dimensions.
}
\scalebox{0.6}{\label{tab:summary_ablation_table_strategy}
\begin{tabular}{llllllll}
\hline
Strategy     & Dimension         & Calibrator & ECE                        & GECE                    & Brier                      & Slope                      & Accuracy                   \\ \hline
DS-Prompt         & Completeness (Ind) & None       & 0.130 $\pm$ 0.036          & 0.114 $\pm$ 0.042          & 0.119 $\pm$ 0.047          & 0.495 $\pm$ 0.207          & 0.699 $\pm$ 0.110          \\
DS-Prompt         & Completeness (Ind) & QAB        & \textbf{0.057 $\pm$ 0.025} & \textbf{0.046 $\pm$ 0.011} & 0.089 $\pm$ 0.031          & \textbf{0.780 $\pm$ 0.180} & 0.644 $\pm$ 0.141          \\
DS-Prompt         & Completeness (Ind) & GIRB       & 0.070 $\pm$ 0.014          & 0.060 $\pm$ 0.023          & \textbf{0.085 $\pm$ 0.024} & 0.750 $\pm$ 0.310          & \textbf{0.681 $\pm$ 0.128} \\ \hline
DS-Prompt         & Conciseness (Ind) & None       & 0.123 $\pm$ 0.027          & 0.127 $\pm$ 0.024          & 0.120 $\pm$ 0.036          & 0.414 $\pm$ 0.208          & \textbf{0.714 $\pm$ 0.103} \\
DS-Prompt         & Conciseness (Ind) & QAB        & \textbf{0.070 $\pm$ 0.027} & \textbf{0.071 $\pm$ 0.014} & 0.102 $\pm$ 0.042          & \textbf{0.584 $\pm$ 0.165} & 0.594 $\pm$ 0.197          \\
DS-Prompt         & Conciseness (Ind) & GIRB       & 0.072 $\pm$ 0.024          & 0.073 $\pm$ 0.025          & \textbf{0.098 $\pm$ 0.022} & 0.567 $\pm$ 0.157          & 0.680 $\pm$ 0.183          \\ \hline
DS-Prompt         & Faithfulness (Ind) & None       & 0.058 $\pm$ 0.020          & 0.097 $\pm$ 0.047          & \textbf{0.101 $\pm$ 0.031} & 0.704 $\pm$ 0.313          & 0.646 $\pm$ 0.097          \\
\hline
\end{tabular}}
\end{table*}

\subsection{Bar Plot Figures}\label{app:figsum}
Figure~\ref{fig:barplot_sum_2} provides a comparison of different calibration methods across different dimensions for various evaluation metrics.
\begin{figure*}[h]
    \centering
    \includegraphics[width=400pt]{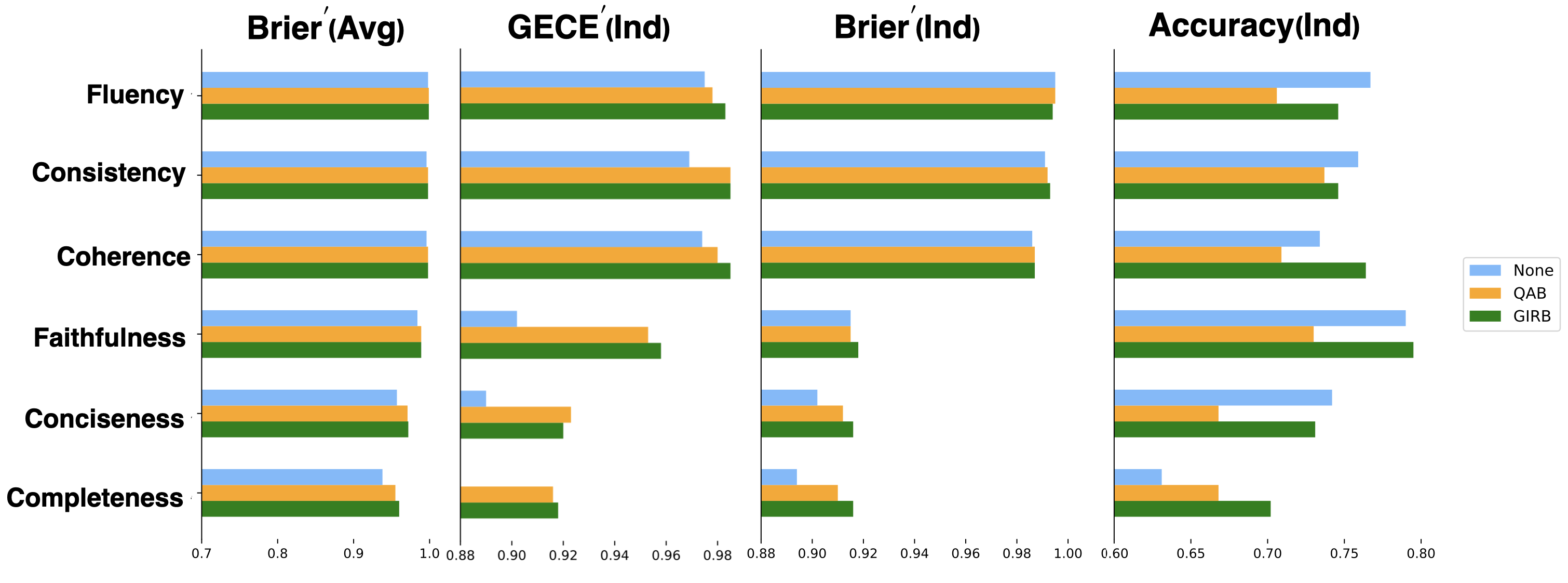}
    \caption{
    Bar Plot Comparison of Calibration Methods Across Dimensions and Metrics
    }
    \label{fig:barplot_sum_2}
\end{figure*}

\section{Question and Answering (QA) Task}\label{app:qa}
\subsection{Prompt for Obtaining Answer and Confidence in Question Answering task}\label{appendix:prompt_qa}
We present the prompt templates used to elicit answers and confidence scores from the language models in our experiments. We employ two prompting strategies to extract confidence estimates alongside model responses: \texttt{verb1s-topk} and \texttt{ling1s-topk}.
\paragraph{\texttt{verb1s-topk}}
{\small \textit{
\\Provide your best guess and the probability that it is correct (from 0.0 to 1.0) for the following question. Give only the guess and probability — no other words or explanations. \\
Example: Guess: <most likely guess, as short as possible; not a complete sentence, just the guess!> \\
Probability: <a number between 0.0 and 1.0 representing how likely your guess is correct, with no extra commentary>. \\
The question is: \textcolor{blue}{<q>}.
}}
\paragraph{\texttt{ling1s-topk}}
{\small \textit{
\\Provide your best guess for the following question, and indicate how confident you are using one of the following expressions: \$EXPRESSION\_LIST. Give only the guess and confidence — no other words or explanations. \\
Example: Guess: <most likely guess, as short as possible; not a complete sentence, just the guess!> \\
Confidence: <one short expression from the list, with no extra commentary>. \\
The question is: \textcolor{blue}{<q>}.
}}

\subsection{Summary Results Tables}\label{app:summarytabqa}
Tables \ref{tab:calibrator_summary_maqa}, \ref{tab:calibrator_summary_OpenBookQA}, \ref{tab:calibrator_summary_SciQ}, \ref{tab:calibrator_summary_TriviaQA}, and \ref{tab:calibrator_summary_TruthfulQA} present comprehensive summaries of calibration method performance across the MathQA, OpenBookQA, SciQ, TriviaQA, and TruthfulQA datasets, respectively. Each table reports two key performance indicators for every calibration metric: ``Gain'', which measures the average improvement relative to the uncalibrated baseline (``None''), and ``Wins'', which counts the number of dimensions in which a given calibration method achieves the best performance among all compared approaches. These summaries enable systematic comparison of calibration techniques across different datasets and identify which methods consistently deliver superior calibration performance. We observe that GIRB either outperforms other calibration approaches or performs comparably in terms of both ``Wins'' and ``Gain'' across all the datasets. 


\begin{table*}[h]
\centering
\caption{Summary of performance across calibration methods for each metric in MathQA dataset. For each metric, we report two indicators: i. Gain: average improvement compared to ``None'', and ii. Wins: the number of setting where a given method outperforms the others.}
\scalebox{0.7}{\label{tab:calibrator_summary_maqa}
\begin{tabular}{lcclclclclclclclc}
\hline
                       & \multicolumn{2}{c}{None} & \multicolumn{2}{c}{S-B}         & \multicolumn{2}{c}{S}                 & \multicolumn{2}{c}{B}           & \multicolumn{2}{c}{QAB}         & \multicolumn{2}{c}{HS-QAB}      & \multicolumn{2}{c}{S-QAB}             & \multicolumn{2}{c}{GIRB}              \\
Metric                 & Gain        & Wins       & \multicolumn{1}{c}{Gain} & Wins & \multicolumn{1}{c}{Gain} & Wins       & \multicolumn{1}{c}{Gain} & Wins & \multicolumn{1}{c}{Gain} & Wins & \multicolumn{1}{c}{Gain} & Wins & \multicolumn{1}{c}{Gain} & Wins       & \multicolumn{1}{c}{Gain} & Wins       \\ \hline
\ECEc   & 0           & 0          & -0.061                   & 0    & \textbf{+0.344}          & \textbf{4} & -0.014                   & 0    & +0.145                   & 0    & +0.017                   & 0    & +0.059                   & 0          & +0.185                   & 0          \\
\GECEc  & 0           & 0          & +0.281                   & 0    & +0.590                   & 0          & +0.329                   & 0    & +0.696                   & 0    & +0.747                   & 0    & \textbf{+0.828}          & \textbf{3} & +0.749          & \textbf{1} \\
\Brierc & 0           & 0          & +1.038                   & 0    & +1.190          & \textbf{2} & +1.116                   & 0    & +1.134                   & 0    & +1.129                   & 0    & +1.144                   & 0          & \textbf{+1.173}          & \textbf{2} \\
AUAC                   & 0           & 0          & -0.230                   & 0    & -0.200                   & 0          & +0.017                   & 0    & +0.036                   & 0    & +0.009                   & 0    & -0.003                   & 0          & \textbf{+0.043}          & \textbf{4} \\ \hline
\end{tabular}}
\end{table*}



\begin{table*}[h]
\centering
\caption{Summary of performance across calibration methods for each metric in OpenBookQA dataset. For each metric, we report two indicators: i. Gain: average improvement compared to ``None'', and ii. Wins: the number of setting where a given method outperforms the others.}
\scalebox{0.7}{\label{tab:calibrator_summary_OpenBookQA}
\begin{tabular}{lcclclclclclclclc}
\hline
                       & \multicolumn{2}{c}{None} & \multicolumn{2}{c}{S-B}         & \multicolumn{2}{c}{S}                 & \multicolumn{2}{c}{B}           & \multicolumn{2}{c}{QAB}         & \multicolumn{2}{c}{HS-QAB}      & \multicolumn{2}{c}{S-QAB}             & \multicolumn{2}{c}{GIRB}              \\
Metric                 & Gain        & Wins       & \multicolumn{1}{c}{Gain} & Wins & \multicolumn{1}{c}{Gain} & Wins       & \multicolumn{1}{c}{Gain} & Wins & \multicolumn{1}{c}{Gain} & Wins & \multicolumn{1}{c}{Gain} & Wins & \multicolumn{1}{c}{Gain} & Wins       & \multicolumn{1}{c}{Gain} & Wins       \\ \hline
\ECEc   & 0           & 0          & +0.388                & 0    & \textbf{+1.085}      & \textbf{4} & +0.532                & 0    & +0.605                & 0    & +0.447                & 0    & +0.483                & 0          & +0.745                   & 0          \\
\GECEc  & 0           & 0          & +0.260                & 0    & +0.759                & 1          & +0.341                & 0    & +0.645                & 0    & +0.642                & 0    & \textbf{+0.754 }      & \textbf{2} & +0.728          & 1 \\
\Brierc & 0           & 0          & +0.431                & 0    & \textbf{+0.636 }      & 1 & +0.603                & 0    & +0.591                & 0    & +0.400                & 0    & +0.413                & 0          & \textbf{+0.636}          & \textbf{3} \\
AUAC                   & 0           & 1          & -0.523                & 0    & -0.657                & 0          & -0.179                & 0    & -0.041                & 0    & -0.053                & 0    & -0.053              & 1          & \textbf{+0.030}          & \textbf{2} \\ \hline

\end{tabular}}
\end{table*}



\begin{table*}[h]
\centering
\caption{Summary of performance across calibration methods for each metric in SciQ dataset. For each metric, we report two indicators: i. Gain: average improvement compared to ``None'', and ii. Wins: the number of setting where a given method outperforms the others.}
\scalebox{0.7}{\label{tab:calibrator_summary_SciQ}
\begin{tabular}{lcclclclclclclclc}
\hline
                       & \multicolumn{2}{c}{None} & \multicolumn{2}{c}{S-B}         & \multicolumn{2}{c}{S}                 & \multicolumn{2}{c}{B}           & \multicolumn{2}{c}{QAB}         & \multicolumn{2}{c}{HS-QAB}      & \multicolumn{2}{c}{S-QAB}             & \multicolumn{2}{c}{GIRB}              \\
Metric                 & Gain        & Wins       & \multicolumn{1}{c}{Gain} & Wins & \multicolumn{1}{c}{Gain} & Wins       & \multicolumn{1}{c}{Gain} & Wins & \multicolumn{1}{c}{Gain} & Wins & \multicolumn{1}{c}{Gain} & Wins & \multicolumn{1}{c}{Gain} & Wins       & \multicolumn{1}{c}{Gain} & Wins       \\ \hline
\ECEc   & 0           & 0          & +0.748                & 0    & \textbf{+1.588 }      & \textbf{4} & +0.901                & 0    & +1.090                & 0    & +0.776                & 0    & +0.837                & 0          & +1.164                & 0          \\
\GECEc  & 0           & 0          & +0.567                & 0    & \textbf{+1.085 }      & \textbf{2}          & +0.648                & 0    & +0.990                & 1    & +0.916                & 0    & +1.027       & 1 & +1.006       & 0 \\
\Brierc & 0           & 0          & +0.777                & 0    & +0.995       & 0 & +0.937                & 0    & +0.936                & 0    & +0.693                & 0    & +0.693                & 0          & \textbf{+1.009 }      & \textbf{4} \\
AUAC                   & 0           & 0          & -0.610                & 0    & -0.660                & 0          & -0.039                & 0    & +0.068                & 0    & +0.024                & 1    & +0.015                & 0          & \textbf{+0.127 }      & \textbf{3} \\ \hline

\end{tabular}}
\end{table*}

\begin{table*}[h]
\centering
\caption{Summary of performance across calibration methods for each metric in TriviaQA dataset. For each metric, we report two indicators: i. Gain: average improvement compared to ``None'', and ii. Wins: the number of setting where a given method outperforms the others.}
\scalebox{0.7}{\label{tab:calibrator_summary_TriviaQA}
\begin{tabular}{lcccccccccccccccc}
\hline
& \multicolumn{2}{c}{None} & \multicolumn{2}{c}{S-B} & \multicolumn{2}{c}{S} & \multicolumn{2}{c}{B} & \multicolumn{2}{c}{QAB} & \multicolumn{2}{c}{HS-QAB} & \multicolumn{2}{c}{S-QAB} & \multicolumn{2}{c}{GIRB} \\
Metric & Gain & Wins & Gain & Wins & Gain & Wins & Gain & Wins & Gain & Wins & Gain & Wins & Gain & Wins & Gain & Wins \\ \hline
\ECEc &0    & 0  & +1.567& 0 & \textbf{+2.718}& \textbf{4} & +2.127& 0 & +2.262& 0 & +1.488& 0 & +1.510 & 0 & +2.317& 0          \\
\GECEc &0 & 0  & +1.541& 0 & \textbf{+2.263} & \textbf{2} & +2.035& 0 & +2.206& 1 & +1.675& 0 & +1.696 & 0 & +2.216& 1          \\
\Brierc &0   & 0  & +1.397& 0 & \textbf{+1.802} & \textbf{2} & +1.759& 0 & +1.744& 0 & +1.225& 0 & +1.214& 0 & +1.763& \textbf{2}          \\
AUAC &0   & 0  & -0.668& 0 & -0.824& 0          & -0.220& 0 & +0.193& 0 & +0.002& 0 & +0.008& 0 & \textbf{+0.345} &\textbf{4} \\ \hline
\end{tabular}}
\end{table*}

\begin{table*}[h]
\centering
\caption{Summary of performance across calibration methods for each metric in TruthfulQA dataset. For each metric, we report two indicators: i. Gain: average improvement compared to ``None'', and ii. Wins: the number of settings where a given method outperforms the others.}
\scalebox{0.7}{\label{tab:calibrator_summary_TruthfulQA}
\begin{tabular}{lcccccccccccccccc}
\hline
& \multicolumn{2}{c}{None} & \multicolumn{2}{c}{S-B} & \multicolumn{2}{c}{S} & \multicolumn{2}{c}{B} & \multicolumn{2}{c}{QAB} & \multicolumn{2}{c}{HS-QAB} & \multicolumn{2}{c}{S-QAB} & \multicolumn{2}{c}{GIRB} \\
Metric & Gain & Wins & Gain & Wins & Gain & Wins & Gain & Wins & Gain & Wins & Gain & Wins & Gain & Wins & Gain & Wins \\ \hline
\ECEc  & 0  & 0  & +1.008 & 0 & \textbf{+2.132} &{\textbf{4}} & +1.936 & 0 & +1.901 & 0 & +0.974 & 0 & +0.924 & 0 & +1.896 & 0          \\
\GECEc & 0 & 0  & +0.981 & 0 & \textbf{+2.073} &{\textbf{3}} & +2.014 & 0 & +1.984 & 0 & +0.994 & 0 & +0.954 & 0 & +2.010 & 1          \\
\Brierc & 0 & 0  & +0.886 & 0 & \textbf{+1.480} &\textbf{2} & +1.464 & 0 & +1.416 & 0 & +0.796 & 0 & +0.776 & 0 & +1.433 &2          \\
AUAC  & 0 & 1  & -0.554 & 0 & -0.909 & 0          & -0.651 & 0 & +0.110 &1  & +0.053 &1 & +0.076 &0 & \textbf{+0.226} &\textbf{1} \\ \hline

\end{tabular}}
\end{table*}

\subsection{Complete Results Tables}\label{app:fulltabqa}
Tables \ref{tab:maqa}, \ref{tab:MMLU}, \ref{tab:OpenBookQA}, \ref{tab:SciQ}, \ref{tab:TriviaQA}, and \ref{tab:TruthfulQA} provide detailed results of calibration method performance across various experimental settings for the MathQA, MMLU, OpenBookQA, SciQ, TriviaQA, and TruthfulQA datasets, respectively. 
Currently, there is no systematic exploration of isotonic regression models on QA datasets. To address this gap, we contribute an ablation study that investigates different variants of isotonic regression, including a version without group-level calibration (IRB). Moreover, inspired by the logic of S-QAB and HS-QAB in \citet{manggala2024qa}, we derive the corresponding group-based versions, S-GIRB and HS-GIRB.
These tables present the full range of performance metrics for each calibration approach under different configurations (LLM, prompting strategies), offering a comprehensive view of method effectiveness and variability. Together, they allow for in-depth comparison and analysis of calibration strategies across multiple benchmarks.

\begin{table*}[h]
\centering
\caption{Performance of different calibration methods across various setting in MathQA dataset.
}
\scalebox{0.8}{\label{tab:maqa}
\begin{tabular}{llllllll}
\hline
Dataset & Prompt      & LLM     & Calibrator & ECE                        & GECE                         & Brier                      & AUAC                       \\ \hline
MathQA  & ling1s-topk & gemma   & QAB        & $0.223 \pm 0.01$           & $0.298 \pm 0.006$          & $0.247 \pm 0.006$          & $0.636 \pm 0.009$          \\
MathQA  & ling1s-topk & gemma   & GIRB       & $0.187 \pm 0.005$          & $0.264 \pm 0.012$          & $0.252 \pm 0.006$          & $\mathbf{0.639 \pm 0.007}$ \\
MathQA  & ling1s-topk & gemma   & HS-QAB     & $0.313 \pm 0.081$          & $0.264 \pm 0.013$          & $0.31 \pm 0.071$           & $0.628 \pm 0.008$          \\
MathQA  & ling1s-topk & gemma   & HS-GIRB    & $0.283 \pm 0.09$           & $0.228 \pm 0.013$          & $0.31 \pm 0.071$           & $0.628 \pm 0.008$          \\
MathQA  & ling1s-topk & gemma   & S-QAB      & $0.24 \pm 0.188$           & $0.218 \pm 0.02$           & $0.317 \pm 0.113$          & $0.613 \pm 0.013$          \\
MathQA  & ling1s-topk & gemma   & S-GIRB     & $0.223 \pm 0.203$          & $\mathbf{0.173 \pm 0.023}$ & $0.317 \pm 0.113$          & $0.613 \pm 0.013$          \\
MathQA  & ling1s-topk & gemma   & S-B        & $0.381 \pm 0.116$          & $0.461 \pm 0.018$          & $0.32 \pm 0.118$           & $0.44 \pm 0.024$           \\
\hline
\end{tabular}}
\end{table*}

\begin{table*}[h]
\centering
\caption{Performance of different calibration methods across various setting in MMLU dataset.}
\scalebox{0.8}{\label{tab:MMLU}
\begin{tabular}{llllllll}
\hline
Dataset & Prompt      & LLM     & Calibrator & ECE                        & GECE                         & Brier                      & AUAC                       \\ \hline
MMLU    & ling1s-topk & gemma   & QAB        & $0.191 \pm 0.006$          & $0.226 \pm 0.006$          & $0.183 \pm 0.01$           & $\mathbf{0.311 \pm 0.039}$ \\
MMLU    & ling1s-topk & gemma   & GIRB       & $0.164 \pm 0.013$          & $0.221 \pm 0.013$          & $\mathbf{0.171 \pm 0.012}$ & $\mathbf{0.311 \pm 0.028}$ \\
MMLU    & ling1s-topk & gemma   & HS-QAB     & $0.336 \pm 0.105$          & $0.308 \pm 0.012$          & $0.299 \pm 0.094$          & $0.206 \pm 0.011$          \\
MMLU    & ling1s-topk & gemma   & HS-GIRB    & $0.328 \pm 0.104$          & $0.298 \pm 0.013$          & $0.299 \pm 0.095$          & $0.206 \pm 0.011$          \\
MMLU    & ling1s-topk & gemma   & S-QAB      & $0.284 \pm 0.17$           & $0.285 \pm 0.02$           & $0.303 \pm 0.135$          & $0.205 \pm 0.018$          \\
MMLU    & ling1s-topk & gemma   & S-GIRB     & $0.277 \pm 0.173$          & $0.275 \pm 0.023$          & $0.303 \pm 0.135$          & $0.206 \pm 0.018$          \\
MMLU    & ling1s-topk & gemma   & S-B        & $0.334 \pm 0.154$          & $0.392 \pm 0.033$          & $0.266 \pm 0.151$          & $0.074 \pm 0.013$          \\
\hline
\end{tabular}}
\end{table*}

\begin{table*}[h]
\centering
\caption{Performance of different calibration methods across various setting in OpenBookQA dataset.}
\scalebox{0.8}{\label{tab:OpenBookQA}
\begin{tabular}{llllllll}
\hline
Dataset    & Prompt      & LLM     & Calibrator & ECE                        & GECE                         & Brier                      & AUAC                       \\ \hline
OpenBookQA & ling1s-topk & gemma   & QAB        & $0.267 \pm 0.011$          & $0.283 \pm 0.011$          & $0.246 \pm 0.014$          & $0.302 \pm 0.031$          \\
OpenBookQA & ling1s-topk & gemma   & GIRB       & $0.209 \pm 0.026$          & $0.255 \pm 0.01$           & $\mathbf{0.215 \pm 0.011}$ & $0.313 \pm 0.041$          \\
OpenBookQA & ling1s-topk & gemma   & HS-QAB     & $0.34 \pm 0.106$           & $0.277 \pm 0.029$          & $0.337 \pm 0.095$          & $0.299 \pm 0.027$          \\
OpenBookQA & ling1s-topk & gemma   & HS-GIRB    & $0.322 \pm 0.114$          & $0.256 \pm 0.03$           & $0.338 \pm 0.095$          & $0.3 \pm 0.027$            \\
OpenBookQA & ling1s-topk & gemma   & S-QAB      & $0.297 \pm 0.161$          & $0.233 \pm 0.032$          & $0.326 \pm 0.127$          & $0.324 \pm 0.036$          \\
OpenBookQA & ling1s-topk & gemma   & S-GIRB     & $0.281 \pm 0.172$          & $0.201 \pm 0.041$          & $0.326 \pm 0.128$          & $\mathbf{0.325 \pm 0.036}$ \\
OpenBookQA & ling1s-topk & gemma   & S-B        & $0.37 \pm 0.122$           & $0.457 \pm 0.03$           & $0.321 \pm 0.129$          & $0.136 \pm 0.028$          \\
\hline
\end{tabular}}
\end{table*}

\begin{table*}[h]
\centering
\caption{Performance of different calibration methods across various setting in SciQ dataset.}
\scalebox{0.8}{\label{tab:SciQ}
\begin{tabular}{llllllll}
\hline
Dataset & Prompt      & LLM     & Calibrator & ECE                        & GECE                         & Brier                      & AUAC                       \\ \hline
SciQ    & ling1s-topk & gemma   & QAB        & $0.23 \pm 0.007$           & $0.281 \pm 0.006$          & $0.218 \pm 0.005$          & $0.253 \pm 0.029$          \\
SciQ    & ling1s-topk & gemma   & GIRB       & $0.177 \pm 0.007$          & $0.237 \pm 0.009$          & $\mathbf{0.195 \pm 0.006}$ & $0.28 \pm 0.013$           \\
SciQ    & ling1s-topk & gemma   & HS-QAB     & $0.315 \pm 0.113$          & $0.27 \pm 0.022$           & $0.301 \pm 0.1$            & $\mathbf{0.286 \pm 0.03}$  \\
SciQ    & ling1s-topk & gemma   & HS-GIRB    & $0.284 \pm 0.12$           & $0.229 \pm 0.022$          & $0.301 \pm 0.1$            & $0.286 \pm 0.03$           \\
SciQ    & ling1s-topk & gemma   & S-QAB      & $0.25 \pm 0.186$           & $0.231 \pm 0.023$          & $0.297 \pm 0.138$          & $0.278 \pm 0.043$          \\
SciQ    & ling1s-topk & gemma   & S-GIRB     & $0.234 \pm 0.197$          & $0.177 \pm 0.03$           & $0.297 \pm 0.138$          & $0.277 \pm 0.043$          \\
SciQ    & ling1s-topk & gemma   & S-B        & $0.351 \pm 0.14$           & $0.433 \pm 0.025$          & $0.291 \pm 0.14$           & $0.105 \pm 0.017$          \\
\hline
\end{tabular}}
\end{table*}

\begin{table*}[h]
\centering
\caption{Performance of different calibration methods across various setting in TriviaQA dataset.}
\scalebox{0.8}{\label{tab:TriviaQA}
\begin{tabular}{llllllll}
\hline
Dataset  & Prompt      & LLM     & Calibrator & ECE                        & GECE                         & Brier                      & AUAC                       \\ \hline
TriviaQA & ling1s-topk & gemma   & QAB        & $0.159 \pm 0.004$          & $0.187 \pm 0.007$          & $0.127 \pm 0.007$          & $0.141 \pm 0.041$          \\
TriviaQA & ling1s-topk & gemma   & GIRB       & $0.138 \pm 0.006$          & $0.171 \pm 0.004$          & $\mathbf{0.105 \pm 0.009}$ & $\mathbf{0.156 \pm 0.042}$ \\
TriviaQA & ling1s-topk & gemma   & HS-QAB     & $0.348 \pm 0.176$          & $0.304 \pm 0.039$          & $0.288 \pm 0.159$          & $0.128 \pm 0.009$          \\
TriviaQA & ling1s-topk & gemma   & HS-GIRB    & $0.336 \pm 0.181$          & $0.287 \pm 0.04$           & $0.288 \pm 0.159$          & $0.128 \pm 0.009$          \\
TriviaQA & ling1s-topk & gemma   & S-QAB      & $0.321 \pm 0.224$          & $0.3 \pm 0.053$            & $0.29 \pm 0.195$           & $0.129 \pm 0.011$          \\
TriviaQA & ling1s-topk & gemma   & S-GIRB     & $0.31 \pm 0.232$           & $0.282 \pm 0.059$          & $0.29 \pm 0.195$           & $0.129 \pm 0.011$          \\
TriviaQA & ling1s-topk & gemma   & S-B        & $0.332 \pm 0.22$           & $0.348 \pm 0.066$          & $0.244 \pm 0.215$          & $0.043 \pm 0.01$           \\
\hline
\end{tabular}}
\end{table*}

\begin{table*}[h]
\centering
\caption{Performance of different calibration methods across various setting in TruthfulQA dataset.}
\scalebox{0.8}{\label{tab:TruthfulQA}
\begin{tabular}{llllllll}
\hline
Dataset    & Prompt      & LLM     & Calibrator & ECE                        & GECE                         & Brier                      & AUAC                       \\ \hline
TruthfulQA & ling1s-topk & gemma   & QAB        & $0.12 \pm 0.024$           & $0.146 \pm 0.03$           & $0.092 \pm 0.028$          & $0.045 \pm 0.056$          \\
TruthfulQA & ling1s-topk & gemma   & GIRB       & $0.12 \pm 0.023$           & $0.14 \pm 0.026$           & $0.097 \pm 0.029$          & $0.055 \pm 0.065$          \\
TruthfulQA & ling1s-topk & gemma   & HS-QAB     & $0.463 \pm 0.175$          & $0.48 \pm 0.093$           & $0.374 \pm 0.175$          & $0.062 \pm 0.028$          \\
TruthfulQA & ling1s-topk & gemma   & HS-GIRB    & $0.463 \pm 0.176$          & $0.479 \pm 0.094$          & $0.373 \pm 0.176$          & $0.063 \pm 0.029$          \\
TruthfulQA & ling1s-topk & gemma   & S-QAB      & $0.482 \pm 0.205$          & $0.492 \pm 0.098$          & $0.384 \pm 0.206$          & $0.059 \pm 0.033$          \\
TruthfulQA & ling1s-topk & gemma   & S-GIRB     & $0.481 \pm 0.206$          & $0.493 \pm 0.1$            & $0.383 \pm 0.206$          & $0.059 \pm 0.033$          \\
TruthfulQA & ling1s-topk & gemma   & S-B        & $0.388 \pm 0.283$          & $0.443 \pm 0.094$          & $0.286 \pm 0.277$          & $0.024 \pm 0.008$          \\
\hline
\end{tabular}}
\end{table*}

\subsection{Bar Plot Figures}\label{app:figqa}
In this section, we present the bar plots in Figures~\ref{fig:bar_plot_qa_maqa}, \ref{fig:bar_plot_qa_MMLU}, \ref{fig:bar_plot_qa_OpenBookQA}, \ref{fig:bar_plot_qa_SciQ}, \ref{fig:bar_plot_qa_TriviaQA}, and \ref{fig:bar_plot_qa_TruthfulQA} to show the comparison of the calibration method performance across the MathQA, MMLU, OpenBookQA, SciQ, TriviaQA, and TruthfulQA datasets, respectively.

\begin{figure*}[h]
    \centering
    \includegraphics[width=400pt]{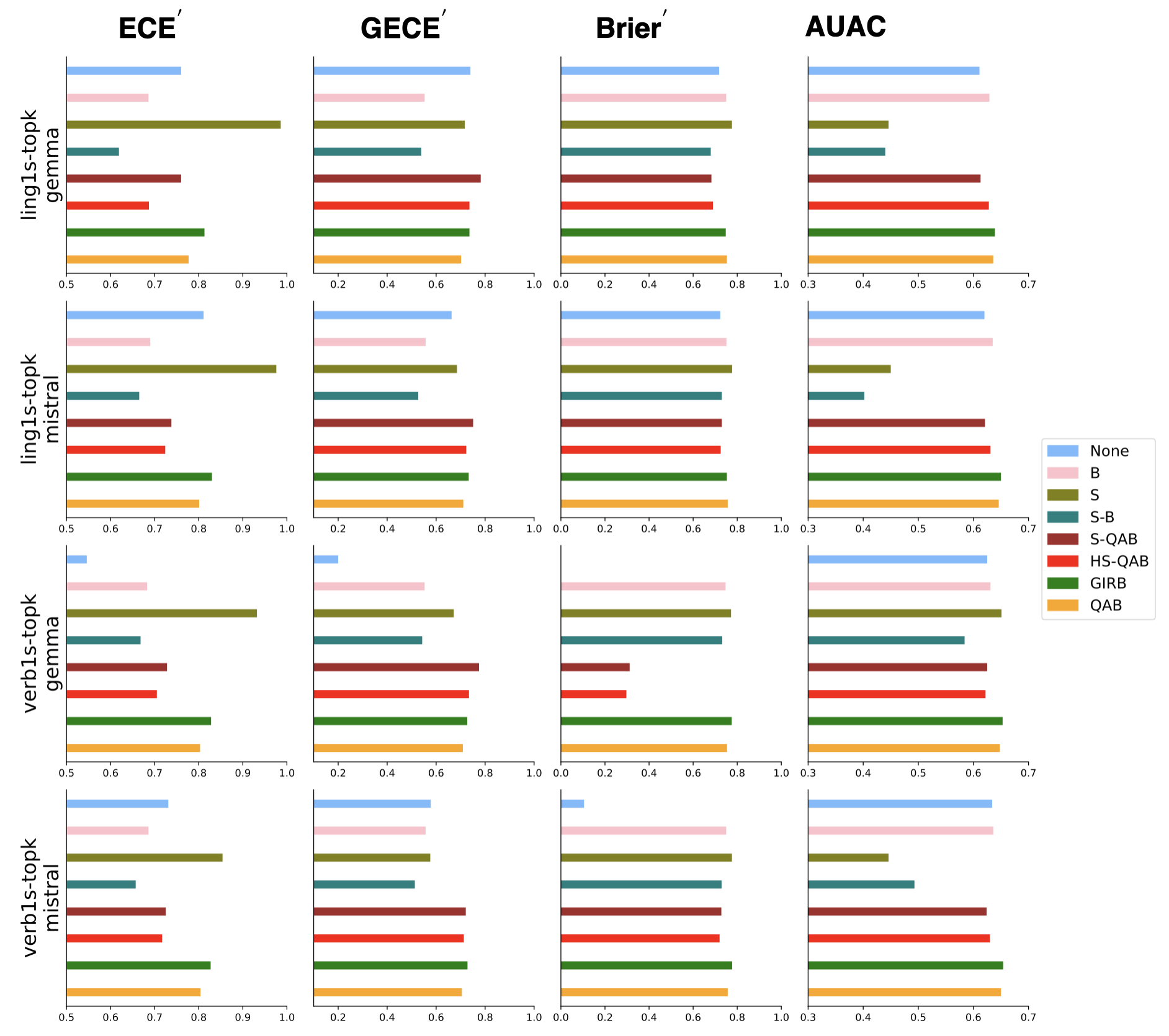}
    \caption{
    Bar Plot Comparison of Calibration Methods Across Settings in MathQA Dataset. 
    }
    \label{fig:bar_plot_qa_maqa}
\end{figure*}

\begin{figure*}[h]
    \centering
    \includegraphics[width=400pt]{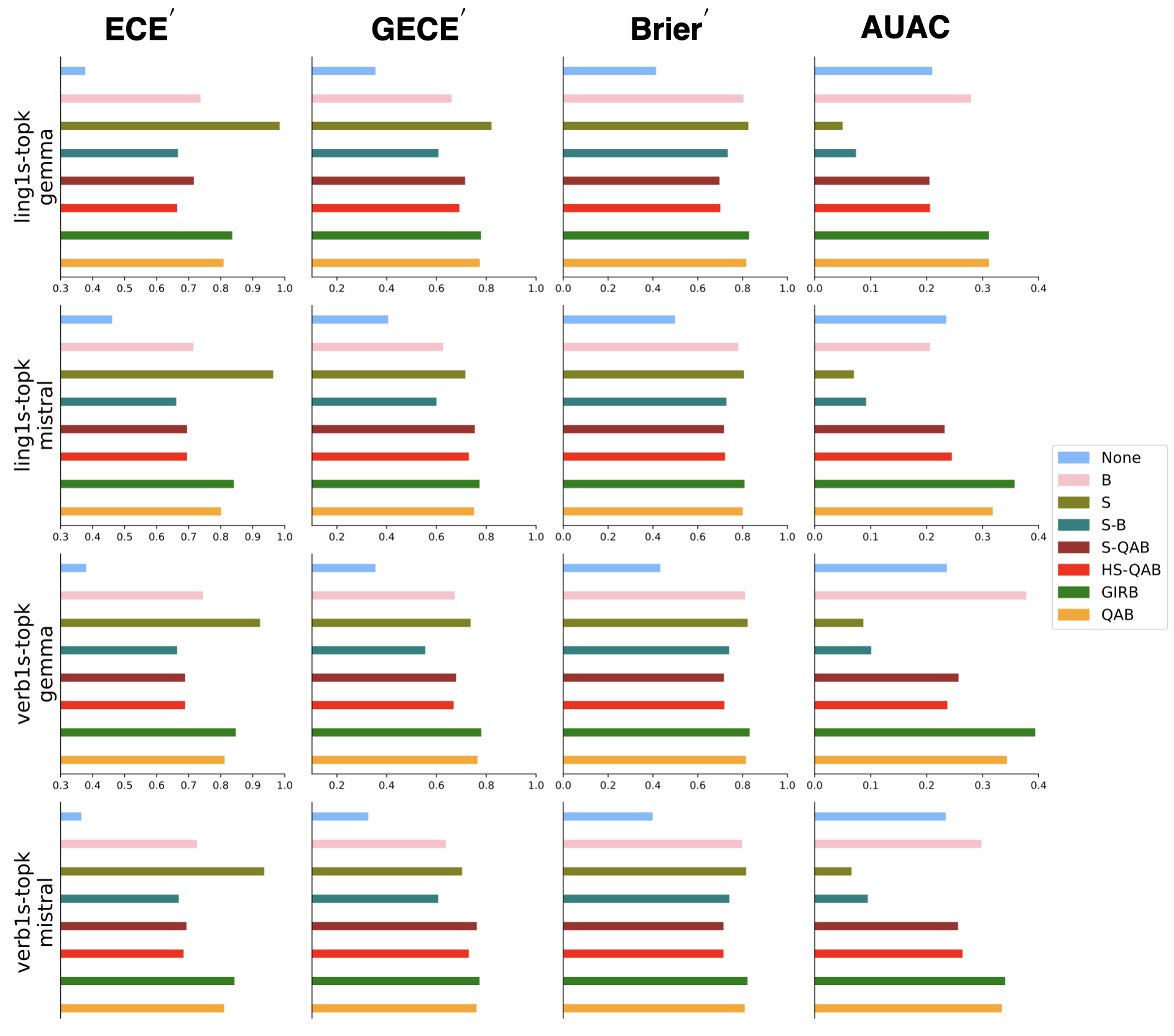}
    \caption{
    Bar Plot Comparison of Calibration Methods Across Settings in MMLU Dataset. 
    }
    \label{fig:bar_plot_qa_MMLU}
\end{figure*}

\begin{figure*}[h]
    \centering
    \includegraphics[width=400pt]{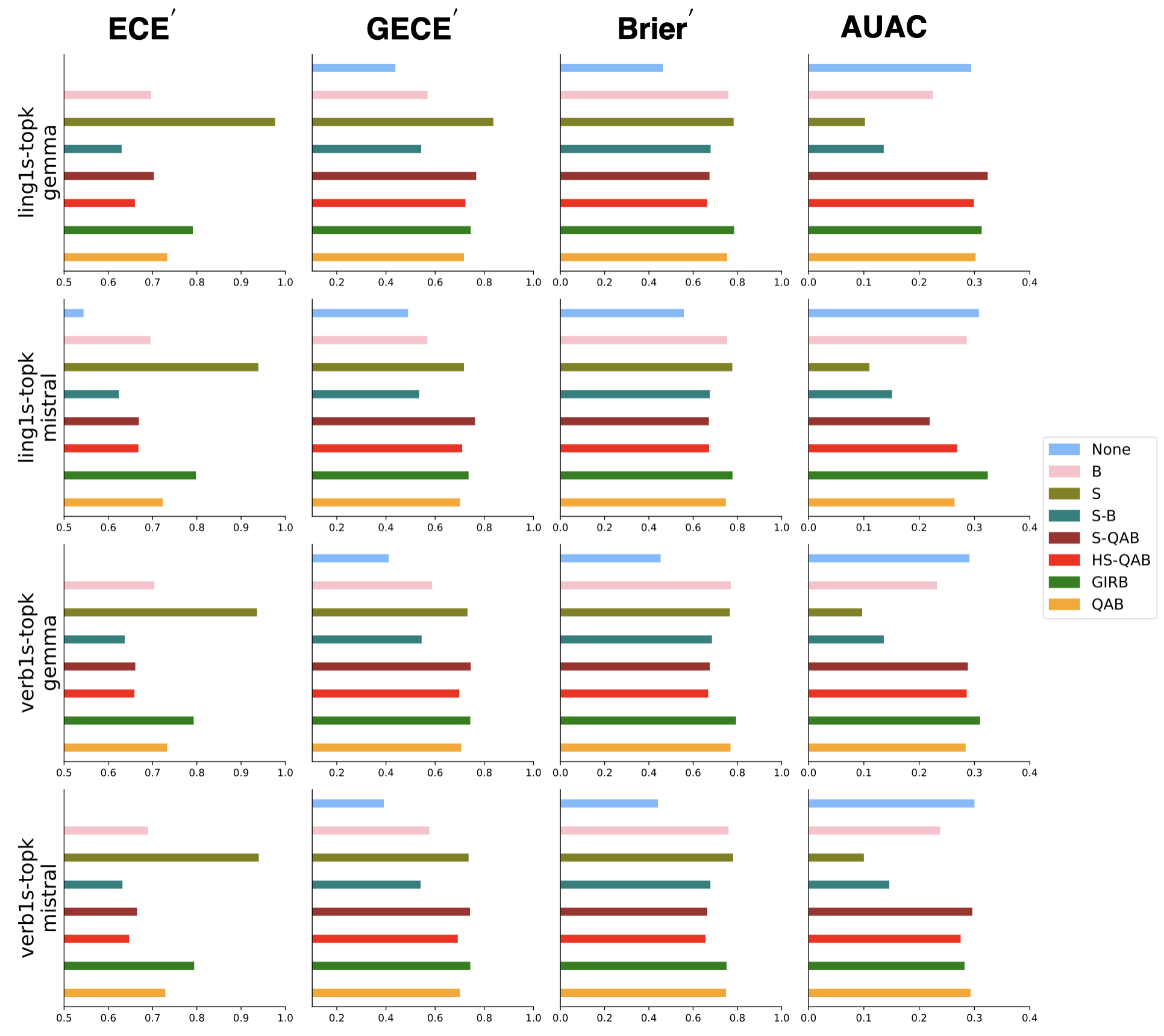}
    \caption{
    Bar Plot Comparison of Calibration Methods Across Settings in OpenBookQA Dataset. 
    }
    \label{fig:bar_plot_qa_OpenBookQA}
\end{figure*}

\begin{figure*}[h]
    \centering
    \includegraphics[width=400pt]{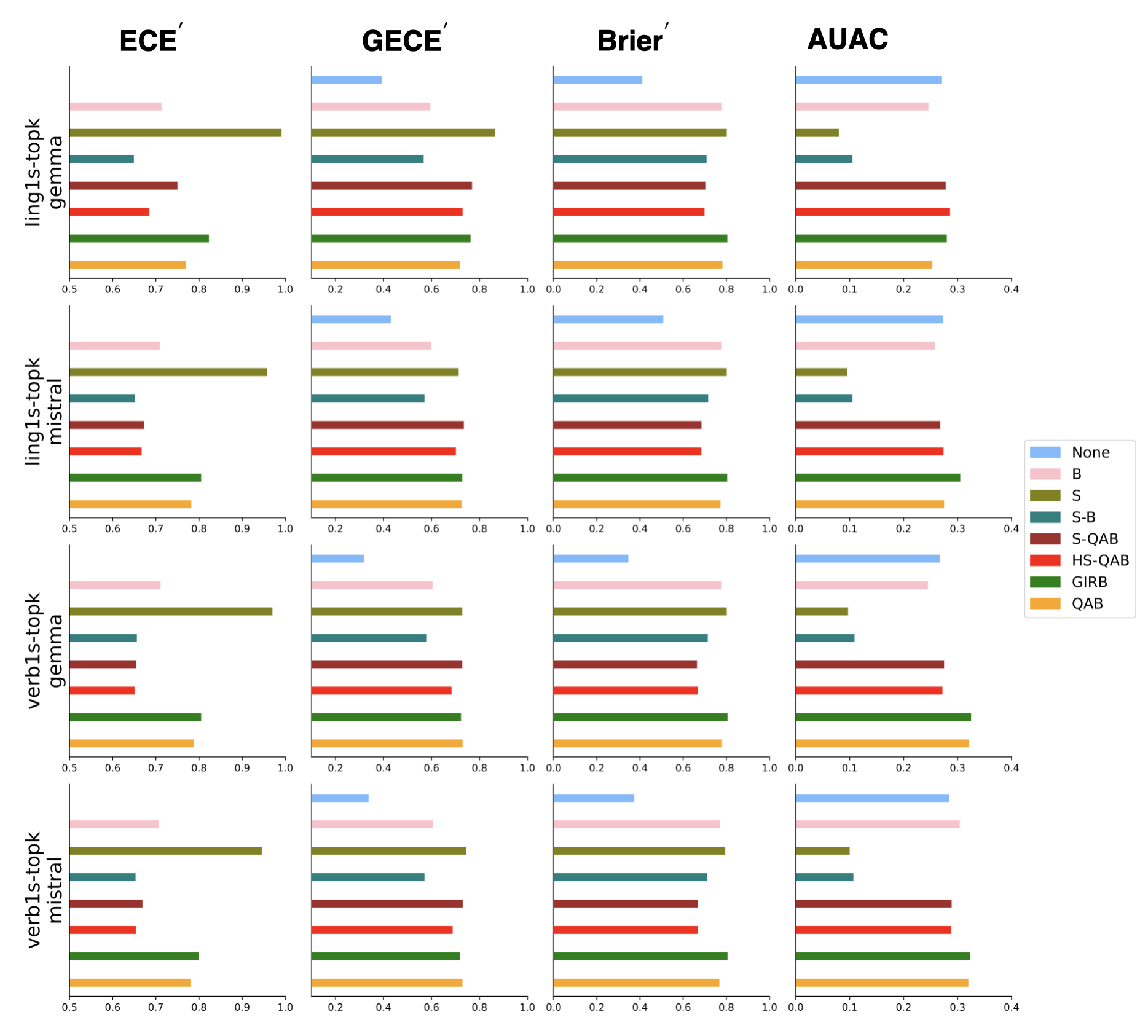}
    \caption{
    Bar Plot Comparison of Calibration Methods Across Settings in SciQ Dataset. 
    }
    \label{fig:bar_plot_qa_SciQ}
\end{figure*}

\begin{figure*}[h]
    \centering
    \includegraphics[width=400pt]{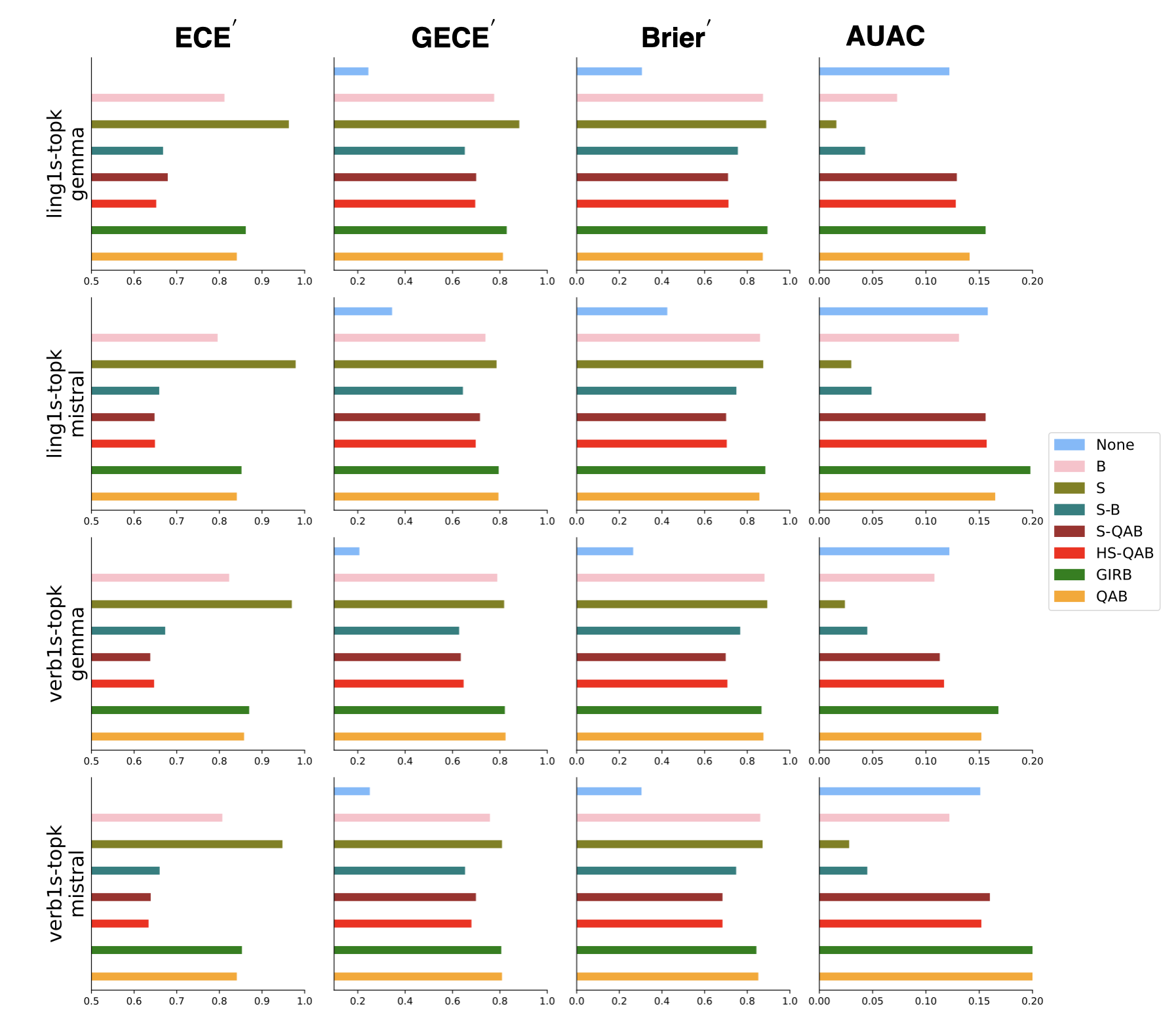}
    \caption{
    Bar Plot Comparison of Calibration Methods Across Settings in TriviaQA Dataset. 
    }
    \label{fig:bar_plot_qa_TriviaQA}
\end{figure*}

\begin{figure*}[h]
    \centering
    \includegraphics[width=400pt]{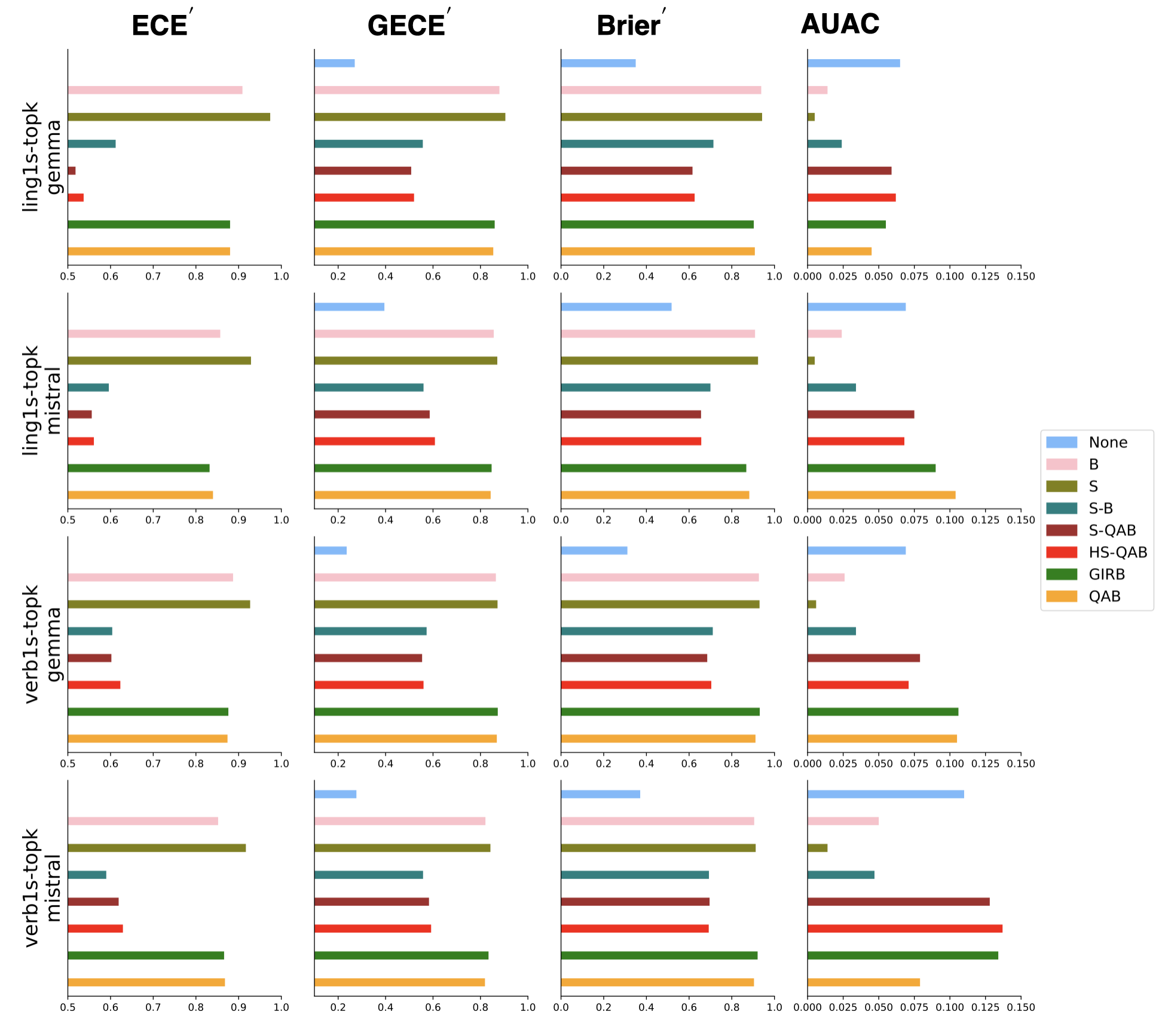}
    \caption{
    Bar Plot Comparison of Calibration Methods Across Settings in TruthfulQA Dataset. 
    }
    \label{fig:bar_plot_qa_TruthfulQA}
\end{figure*}

\end{document}